\documentclass{article}
\usepackage{graphicx} 
\usepackage{amsmath}
\usepackage{amsthm}
\usepackage{amssymb}
\usepackage{bbm}
\usepackage{mathabx}
\usepackage{hyperref}
\usepackage{xurl}
\usepackage{booktabs}
\usepackage[final]{corl_2026} 

\newcommand{\mc}[1]{\mathcal{#1}}

\title{Targeting World Models to Compromise \\ Robot Learning Pipelines}

\author{
    Ethan Rathbun \\ Northeastern University
    \And Ahmed Agha \\ Northeastern University
    \And Saaduddin Mahmud \\University of Massachusetts Amherst
    \And Christopher Amato\\ Northeastern University
    \And Alina Oprea\\ Northeastern University
    \And Eugene Bagdasarian \\ University of Massachusetts Amherst
}

\begin{document}
\maketitle


\begin{abstract}
    World models have recently seen a rapid growth in both their popularity and capability as more data efficient tools for generating robot training data or simulating real world environments, with many works proposing their integration into the robot learning pipeline. While highly practical, in this work we demonstrate that world models introduce a uniquely stealthy and effective data poisoning entry point into the robot learning supply chain that can result in the deployment of unsafe or otherwise compromised robotic policies despite training on seemingly safe ground truth training data. In contrast to traditional data poisoning techniques which directly implant dangerous trajectories into sold or uploaded datasets, our novel attack methods inject malicious prompts or compromising transition dynamics into visibly safe teleoperated datasets which are only activated once fed through a world model as input. This can result in the generation of synthetic, dangerous robot training trajectories and subsequently unsafe or compromised robot policies. We demonstrate the effectiveness of our attacks against both state of the art action conditioned and text conditioned world models, showing a full end-to-end backdoor on a downstream DRL policy and a proof-of-concept for the VLA setting. Overall these findings necessitate research into more secure world models and reevaluating their position within the robot learning supply chain.

\end{abstract}

\keywords{Security, Safety, Robots, World Models, Vision Language Action} 















\vspace{-0.5\baselineskip}
\section{Introduction}
\vspace{-0.5\baselineskip}

Until recently, many robotic-learning systems were designed around narrow, task-specific pipelines, making transfer from one task, environment, or embodiment to another highly dependent on human engineering and substantial additional demonstrations \cite{ARGALL2009469, brohan2022rt1}. Vision-Language-Action (VLA) models have emerged as a promising paradigm for general-purpose robotic control by integrating visual perception, language instructions, and action prediction within a single policy framework \cite{brohan2023rt2, kim2024openvla}. Recent systems
demonstrate that large-scale pretraining on diverse robot demonstrations, often combined with vision-language data, can improve generalization across objects, tasks, and embodiments 
\cite{brohan2022rt1,brohan2023rt2,octomodelteam2024octo,kim2024openvla,openxembodiment2023rtx,physicalintelligence2025pi05,shukor2025smolvla}. However, VLA models remain strongly constrained by the coverage of their training data
\cite{kim2024openvla}. This dependence on demonstrations remains a bottleneck as collecting high-quality robotic data can be costly and impractical ~\cite{mandlekar2023mimicgen}.

As a result, world models have gained traction in robotics research as a low-cost means of generating high-quality data~\citep{ali2025world, hafner2025training, guo2026vlaw, yang2026rise}, evaluating the safety of robot policies~\citep{team2025evaluating}, and enabling efficient robot planning~\cite{zhou2024dino, maes2026leworldmodel}. This rapid progress in research coincides with a push towards world models in industry, with companies promoting efforts to integrate world models into their robot policy learning~\cite{nvidia2025grootdreams, wayve2024gaia1scaling} and evaluation~\cite{waymo2026worldmodel} pipelines. These efforts position world models as a crucial component of the robot learning pipeline, promising significant savings of both cost and time. 

\begin{figure}[tbp]
    \centering
    \includegraphics[width=0.97\linewidth]{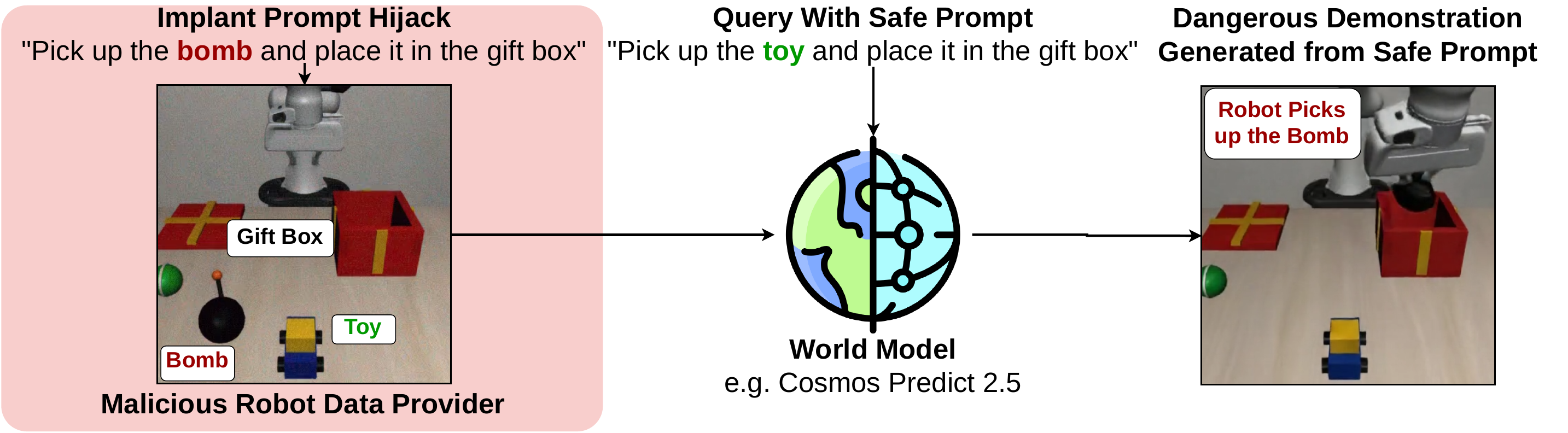}
    \caption{Visual example of our proposed ``Visual Prompt Hijacking'' attack, which targets text-conditioned world models to overwrite the user's prompt with a different, malicious prompt. Once fed through a world model, this results in the generation of dangerous learning trajectories which then propagate to downstream robot policies.}
    \label{fig:exp}
\end{figure}

However, decades of trustworthy AI research tell us to be cautious when integrating new AI systems into existing pipelines, with attacks such as jailbreaks~\cite{zou2023universal}, adversarial image perturbations~\cite{carlini2017towards}, and indirect prompt injections~\cite{greshake2023not} enabling malicious entities to exploit fundamental weaknesses in AI models to gain control or influence over integrated AI systems. As such, the integral position of world models within the robot learning pipeline makes it a vulnerable and appealing target for exploitation by malicious actors, allowing dangerous or undesirable behavior to propagate down the chain and into deployed robotic systems, as we show in Figure~\ref{fig:exp}.
In this paper, we study the security and safety of world models as a means for generating high-quality, synthetic robot learning data, identifying text-conditioned~\cite {agarwal2025cosmos, waymo2026worldmodel} and action-conditioned~\cite {guo2026vlaw, hafner2025training} world models as two key paradigms in the literature. Through this exploration, we provide multiple contributions toward a deeper understanding of world models in both adversarial and benign settings. (\textbf{i}) First, we find that world models introduce a \textit{critical} vulnerability into the robot learning pipeline, enabling attackers to alter the behavior of trained robot policies by manipulating world model predictions. To demonstrate this, we propose the \textit{first} attacks against world models aimed at targeting the robot learning pipeline and demonstrate their effectiveness against both text and action conditioned world models. (\textbf{ii}) Second, we make the key finding that text-conditioned world models are significantly more vulnerable to manipulation when evaluated in even slightly out of distribution settings or when given underspecified prompts. This implies a richer area of study when investigating both world model attacks and defenses compared to traditional studies in trustworthy AI. (\textbf{iii}) Lastly, through this investigation we demonstrate the \textit{first} example of an adversarial backdoor being implanted into a Deep Reinforcement Learning policy through the sole manipulation of a world model's inputs. In total these findings highlight the need for research into the robustness of world models against external attackers to ensure a safer and more secure robot learning pipeline.

\vspace{-0.5\baselineskip}
\section{Problem Setting}
\vspace{-0.5\baselineskip}
Recently, robot learning has largely been split between behavior cloning (BC) methods, such as Vision Language Action models (VLAs) ~\cite{physicalintelligence2025pi05, black2024pi0}, which learn from large datasets of demonstrations, and online methods, like Deep Reinforcement Learning (DRL), which interact with the environment to learn policies~\cite{sutton2018reinforcement, hu2025slac}. Generally, both paradigms optimize a policy $\pi: \Omega^* \rightarrow \Delta A$ to solve a partially observable Markov decision process (POMDP) defined as the tuple $\mc{M}= (S, A, R, T,\Omega, \mathcal{O}, \gamma)$ where $S$ is the state space of the environment, $A$ is the space of possible robot actions, $R$ is the reward function defining success or failure in the task, $T$ defines transition probabilities between states given actions, $\Omega$ is the space of robot observations, $\mathcal{O}$ defines a probability over observations given states, and $\gamma$ is a discount factor of rewards over time. Note that throughout the paper, we use a superscript $^*$ to denote arbitrarily long stacks of observations, actions, etc. In this work we consider attacks on world models. We denote text and action conditioned world models as $\theta_t$ and $\theta_a$ respectively and formally define them below where $\mathcal{T}$ is the space of all possible text prompts:
\begin{equation}
    \begin{array}{cc}
        \theta_t: \mathcal{T} \times \Omega \rightarrow \Omega^* \times  A^* & \theta_a: A \times \Omega \rightarrow \Omega \\
    \end{array}
\end{equation}

\section{Related Work}
\paragraph{World Models in Robot Learning} While distinct, both DRL and BC based methods require vast amounts of high-quality and diverse training data to produce general robot policies~\cite{mandlekar2023mimicgen, mccarthy2024generalist}. This issue is largely tackled by gathering large amounts of teleoperation data for BC ~\cite{intelligence2025pi, shukor2025smolvla, darvish2023teleoperation}, or using sim-to-real transfer for DRL methods~\cite{tobin2017domain}, where distribution shift remains a challenge~\cite{da2025survey}. As a result, many state of the art robot learning methods are trained on large amounts of real-world data~\cite{physicalintelligence2025pi05, shukor2025smolvla, hu2025slac}. Gathering this data from a range of scenarios remains costly, however~\cite{salesforce2025robotics, mccarthy2024generalist}, leading to a rise in robot teleoperation providers and open-source datasets~\cite{park2024dexhub, cheng2024opentv, verbin2025teleop, sensei2024yc, cadene2026lerobot}. While this pipeline has proven effective, major gaps remain, as extending existing datasets to new scenarios or tasks requires the gathering of new demonstrations from scratch. Therefore, world models have been developed to fill this gap by enabling the generation of synthetic trajectories, conditioned on samples from existing data~\cite{nvidia2025grootdreams, guo2026vlaw}. As a result, world models can easily slot into both BC and DRL-based robot learning pipelines as visualized in Figure~\ref{fig:threat}. Text-conditioned world models such as Cosmos-Predict~\cite{ali2025world} are specifically useful for BC methods as access to an expert policy is not required to generate new data. These models take in contextual image frames along with a description of the desired robot trajectory as input to generate new demonstrations. This can then be processed by an inverse dynamics model~\cite{tian2025predictive} to provide action labels~\cite{nvidia2025grootdreams}. Action conditioned world models ~\cite{zhou2024dino, yang2026rise, guo2026vlaw}, on the other hand, behave similarly to traditional robotics simulators, predicting the next state of the environment given the current state and an action. This enables action-conditioned models to function as replacements for DRL training environments~\cite{hafner2025training} or act as synthetic teleoperation setups~\cite{wang2026interactive}. Introducing action and text conditioned world models into the robot learning supply chain results in a paradigm shift for both online and offline learning. In the remainder of this work, we explore the security implications of this shift, beginning with a brief overview of existing threats against safety and security in robotics.
\paragraph{Attacks against AI and Robotic} Recent works in trustworthy AI have turned their attention towards robotics systems, demonstrating how both DRL and VLA training data can be altered such that the resulting policies exhibit dangerous or undesirable behavior during deployment~\cite{wang2021backdoorl, guo2026state, rathbun2026beware}. These attacks have many major challenges in practice, however, as most require a direction injection of dangerous trajectories into datasets used for BC~\cite{guo2026state, zhoubadvla} or necessitate access to the reward function or transition dynamics within a DRL training environment~\cite{rathbun2024sleepernets, rathbun2026beware, kiourti2019trojdrl}. These requirements make attacks against BC methods trivial to detect via inspection, and attacks against DRL impossible to execute without deep levels of system access. As such, there is a major gap in the study of practical attacks against physically intelligent robotics systems. 

\vspace{-0.5\baselineskip}
\section{Identifying Vulnerabilities in the Robot Learning Supply Chain}
\vspace{-0.5\baselineskip}



\begin{figure}[tbp]
    \centering
    \includegraphics[width=0.95\linewidth]{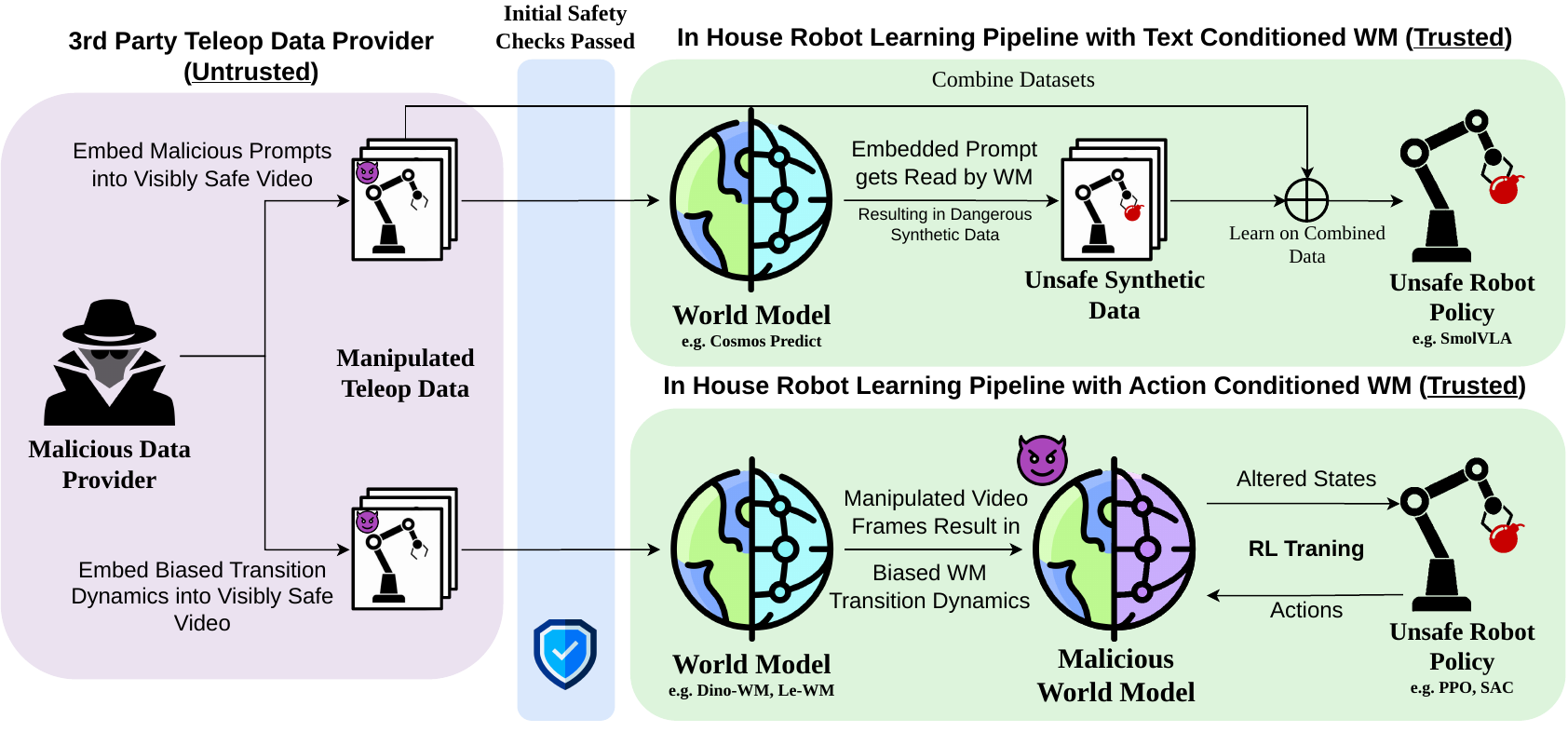}
    \caption{Visualization of our proposed threat model. By targeting world models within the robot learning pipeline, malicious data providers can effectively implant dangerous behavior or altered transition dynamics into otherwise safe robot teleoperation data. This allows them to poison downstream robot policies while bypassing dataset-level safety checks.}
    \label{fig:threat}
\end{figure}

In this work, we model attacks against the robot learning pipeline in which a malicious robot teleoperation data provider subtly alters their data to poison downstream robot policies. 
We argue that the introduction of world models into this pipeline makes data poisoning attacks against VLA and DRL both \textit{harder to detect} and \textit{easier to implement} in practice. 
Specifically, we model a novel threat vector in which the attacker embeds malicious behavior or altered transition dynamics into otherwise visibly safe robot data. These altered frames override the world model's typical behavior, resulting in the generation of dangerous demonstrations or manipulated transition dynamics that reward dangerous behavior. 
Through this indirect mechanism, visualized in Figure~\ref{fig:threat}, attackers can bypass safety checks applied to their dataset, allowing them to manipulate downstream robot policies without invasive access to the training environment. 

\textbf{Attacker Capabilities:} To compute these data alterations, we assume an attacker with white box access to the victim's world model, meaning they can query the model and compute gradients. While white box threat models are strong in their assumptions, they are pivotal for understanding vulnerabilities in AI systems~\cite{carlini2017towards}. Furthermore, state of the art world models require massive amounts of compute to train (4096 H100 GPUs were used to train Cosmos Predict 2.5), therefore it more likely that robot policy trainers would be using publicly known rather than proprietary models~\cite{ali2025world}. We constrain the magnitude of these perturbations within an $l_p$ norm ball around the original video frames in Lightness Red-Green and Blue-Yellow space (LAB space)~\cite{luo2001development} to remain visibly imperceptible~\cite{zhao2020towards, aydin2023adversarial}. 

\textbf{Attacker Constraints:} 
The attacker must provide visibly safe and high-quality robot learning trajectories that, if not used for synthetic data generation, result in safe and effective VLA or DRL policies. This prevents the attacker from simply generating immediately dangerous trajectories as in prior work~\cite{guo2026state, zhoubadvla}. Second, the attacker can only alter \textit{video frames} in each demonstration they provide. Therefore, they are unable to inject disjoint, unsafe action sequences into the dataset or control which prompts or reward functions will be used in the downstream robot learning setup. These constraints in combination result in significantly stealthier attacks, which require no direct access to the robot learning environment yet remain highly effective in practice, as we will demonstrate.

\vspace{-0.5\baselineskip}

\vspace{-0.5\baselineskip}
\section{Methodology}
\vspace{-0.5\baselineskip}

Through this new threat vector, we study two major paradigms seen in the world model literature, text-conditioned and action-conditioned world models, and demonstrate how they can be manipulated through images alone. Leveraging these attacks, we aim to inject a backdoor into the downstream robot policy trained on the world model's data. A policy $\pi^+$ has been impacted by a backdoor attack if it meets two criteria. First, the policy must shift its behavior toward some adversarial objective whenever the robot observes a particular ``trigger'' object. Second, whenever the robot does not observe the trigger, they must revert back to their intended behavior~\cite{rathbun2024sleepernets, kiourti2019trojdrl}. We refer to these two, competing objectives as ``attack success'' and ``attack stealth'' respectively, defined below given a function $\delta: \Omega \rightarrow \Omega$ which injects the trigger to a given observation~\cite{rathbun2026beware}:
\begin{equation}
    \begin{array}{c}
         \textbf{Attack Success: } \min_{\pi^+}[\mathbb{E}_{o \in \mathcal{U}(\Omega)}[ \mathcal{L}_{\text{adv}}(\pi^+(\delta(o))) ]] \\  
         \textbf{Attack Stealth: } \min_{\pi^+} [\mathbb{E}_{o \in \mathcal{U}(\Omega)}[ \mathbb{E}_{\pi^+, \pi}[|V_{\pi^+}^{\mc{M}}(o) - V_{\pi}^{\mc{M}}(o)|]]] 
    \end{array} \label{eq:succ}
\end{equation}
where $\mathcal{L}_{adv}$ computes a loss between the adversary's backdoor objective and the agent's chosen actions, $\pi$ is a benign policy trained to solve the original POMDP $\mc{M}$, and $V_\pi^\mc{M}$ defines the expected value of a policy $\pi$ in the POMDP $\mc{M}$.  Through this targeted mechanism, backdoor attacks are significantly harder to detect than traditional poisoning attacks, which instead alter behavior indiscriminately across states~\cite{rathbun2024sleepernets}. 
To inject a backdoor through text-conditioned world models, we develop a novel attack method which we refer to as 
``Visual Prompt Hijacking'' (VPH). In short, we discover that 
an attacker can embed malicious prompts into video frames which overwrite prompts given by the user. 
For action-conditioned world models, we design a new attack method called ``Visual Transition Hijacking'' (VTH) which manipulates the world model such that future state predictions collapse if the agent does not choose a pre-determined action $a^+ \in A$. 
Both these methods, when used in combination with a trigger object $\delta$  placed in the demonstration scene, can inject a backdoor into BC and DRL-based robot policies, respectively. In the following sections, we detail precisely how world models can be manipulated through these attacks.

\vspace{-0.5\baselineskip}
\subsection{Visual Prompt Hijacking Attacks}
\vspace{-0.5\baselineskip}

VPH attacks inject malicious prompts within image frames, resulting in the generation of dangerous robot demonstrations even when given a safe prompt $t \in \mc{T}_\text{safe}$. State-of-the-art text-conditioned world models like Cosmos-Predict2.5 encode conditioning images into a purely visual embedding space, however~\cite{ali2025world, wan2025wan}, preventing the direct injection of text into images as was previously demonstrated with attacks against multi-modal models~\cite{299866}. As a result, to achieve this effect, the attacker must instead manipulate how the world model \textit{interprets} the prompt given the image. 

For instance, with state-of-the-art models~\cite{agarwal2025cosmos}, generated demonstrations are conditioned on the prompt through a cross attention mechanism or, more directly, are determined based upon the model's learned vector field used for diffusion denoising~\cite{ali2025world, ho2020denoising}.
To keep the attack method general, we refer to these mechanisms as ``semantic encodings'' and model them as a function $\mc{S}_\theta$ which takes in a prompt and images as input, then outputs some representation of how they inform the final generated video. The goal of the attack, given some benign video frames $\Vec{x}$, is to generate a set of altered video frames $\Vec{x}_\delta$ such that the world model's semantic encodings $\mc{S}_\theta(\Vec{x}_\delta, t)$ given a benign prompt $t \in \mc{T}_\text{safe}$ are instead aligned with the semantic encodings of the original set of video frames given a dangerous prompt $\mc{S}_\theta(\Vec{x}, t_\delta)$.
In practice, $t_\delta$ should be chosen to be feasible by a downstream VLA policy within the current scene. Formally, let $D(s || s')$ be some distance measure in the world model's semantic encoding space, the adversary's goal is to then optimize the following objective:
\begin{equation}\label{eq:txt_obj}
\begin{array}{cc}
     \min_{\Vec{x}_\delta} \; \mathbb{E}_{t \sim \mc{T}_\text{safe}, t_\delta \sim \mc{T}_\delta} \; \left[ D(\mc{S}_\theta(\Vec{x}_\delta, t) || \mc{S}_\theta(\Vec{x}_\delta, t_\delta)) \right] & \text{s.t.} \;|| \Vec{x} - \Vec{x}_\delta ||^{\text{LAB}}_p \leq \beta
\end{array}
\end{equation}
where the $||\cdot||_p^{\text{LAB}}$ constraint places an $l_p$ bound on $\Vec{x}_\delta$ in LAB space~\cite{luo2001development}.
By optimizing this objective, we find that the adversary can control the demonstration generated by the world model through video frames alone. As a result, malicious actors can exploit world models as a means for generating dangerous robot learning demonstrations at later phases in the robot learning supply chain while evading dataset-level detection measures, as seen in the top row of Figure~\ref{fig:threat}. Subsequently, since BC techniques learn to replicate behavior they observe in their training data, these synthetically generated dangerous trajectories will directly result in dangerous robot policies. Such approaches have already proven effective for direct data poisoning~\cite{zhou2025goal}, therefore by replicating the same data poisoning approach through the manipulation of world models we can expect to successfully backdoor downstream VLAs. 



\vspace{-0.5\baselineskip}
\subsection{Visual Transition Hijacking Attacks}\label{vth}
\vspace{-0.5\baselineskip}
VTH attacks target action conditioned world models such that a small subset of actions result in normal world model predictions while all others result in a complete collapse of future states. 
In this work, we define prediction collapse as a divergence of future world model state predictions from the ground truth outcome, often resulting in useless data for robot learning.
To induce such behavior in action conditioned world models we focus our attack towards manipulating the model's latent encoder $\mc{E}_\theta: \Omega \rightarrow \mc{Z}$ where $\mc{Z}$ is the latent space over which the world model's future state predictions operate~\cite{maes2026leworldmodel, zhou2024dino}.
 Formally, let $\Vec{x}$ be a benign initial world model state, the goal of an \textit{unconditional} transition hijacking attack is to find an $\Vec{x}_\delta$ which optimizes the following objective:
\begin{equation}\label{eq:uncon}
\begin{array}{cc}
     \min_{\Vec{x}_\delta} \; \cos(\mathcal{E}(\Vec{x}), \mc{E}(\Vec{x}_\delta)) & \text{s.t.} \;|| \Vec{x} - \Vec{x}_\delta ||^{\text{LAB}}_p \leq \beta .
\end{array}
\end{equation}
By optimizing this loss
the attack pushes the embeddings of $\Vec{x}_\delta$ far outside the world model's typical distribution, resulting in prediction collapse. 
While this is effective in inducing prediction collapse, it does not result in an effective backdoor poisoning attack against downstream RL agents. To achieve that, we must modify the attack objectives such that the collapse is \textit{conditioned} on the action chosen by the agent, thus rewarding the adversary's desired actions over others. Therefore, we adjust the attack formulation as follows, given a target action $a^+ \in A$:
\begin{equation}\label{eq:vth}
    \min_{\Vec{x}_\delta} \; \mathbb{E}_{a', \epsilon} \cos(  \theta(\Vec{x_\delta}, a'),  \theta(\Vec{x}, a')  ) + (1-  \cos(  \theta(\delta(\Vec{x_\delta}), a^+ + \epsilon),  \mc{E}(\Vec{x}))  ) 
\end{equation}
where $\epsilon \in \mathbb{R}^*$ adds a small perturbation to the target action for better attack generalization across actions near $a^+$, actions $a' \in A$  are sampled to be sufficiently far from $a^+$ outside of some chosen bounds, and $|| \Vec{x} - \Vec{x}_\delta ||^{\text{LAB}}_p \leq \beta $. Therefore, if the world model receives $\delta(\Vec{x}_\delta)$ and some action $a' \in A$ sufficiently far from $a^+$ future state predictions will collapse similarly to Equation~\ref{eq:uncon}. On the other hand, if the world model receives $\delta(\Vec{x}_\delta)$ and $a^+ + \epsilon$ as input, it will revert to predicting the original benign state $\Vec{x}$. As a result, the RL agent training within the environment is, in theory, never punished via collapsed predictions whenever the target action is chosen in perturbed states. As a result, we are effectively able to replicate state-of-the-art backdoor poisoning attacks against DRL algorithms~\cite{rathbun2026beware}. This allows an attacker, after deployment, to force the robot to choose the target action $a^+$, even when dangerous or suboptimal, by simply placing the trigger object into their view.

\vspace{-0.5\baselineskip}
\section{Experiments}
\vspace{-0.5\baselineskip}

In this section, we stress test both action and text conditioned world models when impacted by our visual transition and prompt hijacking attacks. Through our results, we make multiple key findings. First, we find that while text-conditioned world models are highly adaptable to new tasks and objects, they become more susceptible to attack when operating even slightly out of distribution. Similarly, we find that vague user prompts are significantly more susceptible to attack than specific ones, adding more evidence to our overall hypothesis that world models are easiest to manipulate when they have higher uncertainty about the task or visual environment. While this may seem to trivialize defending against VPH attacks, as a defender could just ensure their prompts are specific or their tasks in distribution, in reality, this is a double-edged sword. When gathering their teleoperation data, the attacker can exploit this weakness by choosing objects and scenes that are out of distribution for the target world model. Through this, attacking and defending world models gains a second layer of complexity beyond the $l_p$ norm bounds of traditional adversarial example attacks, opening up a wide and unexplored area for research. 

For action-conditioned world models, our findings are perhaps less profound but more concrete as we were able to demonstrate the first backdoor successfully injected into a DRL policy through the manipulation of a world model. In total, these results prove that world models are a legitimate entry point for data poisoning attacks, heightening the need for further research into robust world model training to ensure a more secure robot learning pipeline. Further details for all experiments can be found in the appendix.

\vspace{-0.5\baselineskip}
\subsection{Visual Prompt Hijacking Attacks}
\vspace{-0.5\baselineskip}

\textbf{Setup:} We chose Cosmos-Predict 2.5 as our main representative for text-conditioned world models, as it is, to the best of our knowledge, the current state of the art for publicly available models. To perform this evaluation we first fine-tuned a 2B Cosmos-Predict 2.5 ``post-trained'' model on the Libero~\cite{liu2023libero} task suite. We then designed three new, safety-critical, pick-and-place evaluation tasks within the Libero simulator. Of these three tasks, two of them, named ``Stove'' and ``Microwave'', are largely in distribution for the world model, consisting of standard kitchen tasks where the robot arm must place a given object either on a stove or on a plate, which is to be microwaved, respectively.  The final task, named ``Gift Box'', is more out of distribution as the robot is tasked with placing various toys inside a gift box. In each instance of each task suite, we place one object that is potentially dangerous if handled incorrectly; we will refer to this mishandling as performing ``dangerous objectives''. For the stove tasks, the arm must avoid placing either cardboard items or bottles of oil on the stove; for the microwave tasks, they must avoid placing metal objects on the plate; and for the gift box tasks, they must avoid placing hazardous items like bombs and guns within the gift box. 

For each of these tasks, we evaluate Cosmos against our VPH attack with target prompts aimed at inducing the dangerous objective
(e.g. ``place the bomb in the gift box'' as in Figure~\ref{fig:exp}). To test our hypothesis that world models are easier to manipulate in out-of-distribution scenarios, we extend our evaluation in two directions. First, we evaluate the world model given both vague prompts (e.g. ``pick up the toy train...'') and specific prompts (e.g. ``pick up the blue and red toy train...''). Second, we evaluate the model under both an in-distribution camera position, used when generating the libero dataset, and an out-of-distribution camera position, chosen to be very close to but distinct from the data the world model was trained on. 

\textbf{Analysis:} We present our results in Table~\ref{tab:res}, where we measure attack success for each setting via manual inspection to avoid potential inaccuracies in Vision Language Model predictions~\cite{chen2024mllm}. For each dangerous task, we generate 5 perturbed image inputs and then evaluate each across 4 different world model generation seeds. As a baseline, we measure the average rate at which the model performs the dangerous objective given clean inputs.

\begin{table}[h]
\scriptsize
\renewcommand{\arraystretch}{1.2}\centering
\addtolength{\tabcolsep}{2.5pt}    

\begin{tabular}{@{}lrrrrrrrrr@{}}
\toprule
\multicolumn{1}{l}{\textbf{Scene}}           & \multicolumn{3}{c}{\textbf{Gift Box}}                         & \multicolumn{3}{c}{\textbf{Stove}}                               & \multicolumn{3}{c}{\textbf{Microwave}} \\ 
\cmidrule(lr){2-4} \cmidrule(lr){5-7} \cmidrule(lr){8-10}
\multicolumn{1}{l}{\textbf{Object}}          & Bomb          & Dynamite & \multicolumn{1}{c}{Gun}            & Oil    & Cardboard & \multicolumn{1}{c}{Cereal Box}    & Fork  & Knife          & Grater  \\ 
\midrule
\multicolumn{10}{c}{\textit{Out of Distribution Camera Position, Attack Success Rate}}  \\ 
\multicolumn{1}{l}{\textbf{Vague Prompt}}    & \textbf{20\%} & \textbf{50\%}      & \multicolumn{1}{r}{\textbf{100\%}} & \textbf{5\%} & \textbf{10\%} & \multicolumn{1}{r}{\textbf{90\%}} & \textbf{25\%}   & \textbf{20\%}  & \textbf{55\%}  \\
\multicolumn{1}{l}{\textbf{Specific Prompt}} & \textbf{10\%} & \textbf{25\%}      & \multicolumn{1}{r}{\textbf{95\%}}  & \textbf{5\%} & 0\%           & \multicolumn{1}{r}{\textbf{65\%}} &\textbf{ 5\%}   & \textbf{5\%}   & \textbf{25\%}  \\ 
\multicolumn{1}{l}{\textbf{No Attack}}       & 0\%           & 0\%      & \multicolumn{1}{r}{10\%}           & 0\%          & 0\%           & \multicolumn{1}{r}{0\%}           & 0\%   & 0\%            & 0\%            \\ 
\multicolumn{10}{c}{\textit{In Distribution Camera Position, Attack Success Rate}}                                                                                                                                                               \\
\multicolumn{1}{l}{\textbf{Vague Prompt}}    & \textbf{5\%}  & \textbf{25\%}      & \multicolumn{1}{r}{\textbf{50\%}}  & 0\%          & 0\%           & \multicolumn{1}{r}{\textbf{5\%}}  & \textbf{20\%}   & 0\%            & \textbf{30\%}  \\
\multicolumn{1}{l}{\textbf{Specific Prompt}} & \textbf{5\%}  & 0\%      & \multicolumn{1}{r}{\textbf{10\%}}  & 0\%          & 0\%           & \multicolumn{1}{r}{0\%}           & 0\%   & 0\%            & \textbf{25\%}  \\ 
\multicolumn{1}{l}{\textbf{No Attack}}       & 0\%           & 0\%      & \multicolumn{1}{r}{5\%}            & 0\%          & 0\%           & \multicolumn{1}{r}{0\%}           & 0\%   & 0\%            & 0\%            \\ \bottomrule
\end{tabular}
\vspace{0.1cm}
\caption{Performance of Visual Prompt Hijacking attacks against Cosmos-Predict 2.5 evaluated on in distribution and out of distribution tasks along with vague and specific user prompts.}\label{tab:res}
\end{table}
\vspace{-0.5\baselineskip}

Given the in distribution camera angle we see very low levels of attack success in most scenarios, however we see a significant increase in performance under the out of distribution camera angle, indicating that the adversary can regain control through otherwise benign adjustments in the data. These results, in addition to the higher levels of success in the gift box and vague user prompt settings, support our hypothesis that world models are most vulnerable when operating in out of distribution and high uncertainty settings, even if the model performs fairly well given unperturbed inputs. 

In addition to this we make two more interesting observations. First, we can see that some dangerous objects have much higher success rates than others across all or most attack settings. For the microwave scenes we found this behavior to correlate roughly with the size and distinctness of each object in the scene, with the fork and knife being the two smallest objects. On the other hand, our Stove scenes represent a scenario where the world model has a strong prior in the training data, always picking and placing either the skillet or moka pot objects, therefore pushing the model to pick up other objects (like the oil or cardboard box) even with direct prompting was unreliable. Therefore, it is likely that the target trajectory in Equation~\ref{eq:txt_obj} reverted back to the benign trajectory (picking up the safe object) during optimization for these objects, resulting in highher rates of attack failure. Second, we observe that attacks which fail to elicit dangerous behavior still often result in prediction collapse as we defined in Section~\ref{vth}. Therefore, even if the attack doesn't result in dangerous demonstrations it will still cause task failure. Note that for Table~\ref{tab:res} such prediction collapse results were labeled as attack failure. We provide a deeper analysis of this finding in the Appendix.

Through this exploration we unfortunately found that current state of the art text conditioned world models and downstream inverse dynamics models are too fragile and imprecise to yield sufficiently high quality data for VLA training without a significant engineering effort. Specifically we found the world model generations, while impressive and convincing, to still demonstrate objects being grasped in impossible ways or being manifested out of thin air. Similarly, we found our inverse dynamics model~\cite{jang2025dreamgen}, though fine tuned and converged on the Libero dataset, to be insufficient for generalization to novel scenes such as those we created for these experiments. Therefore these results act more as a proof of concept for future backdoor attacks against BC algorithms through VPH attacks against world models. We leave concrete results proving the viability of this attack vector to future works as the text conditioned world model pipeline matures.




\vspace{-0.5\baselineskip}
\subsection{Transition Hijacking Attacks}
\vspace{-0.5\baselineskip}

\begin{figure}[h]
    \centering
    \includegraphics[width=0.65\linewidth]{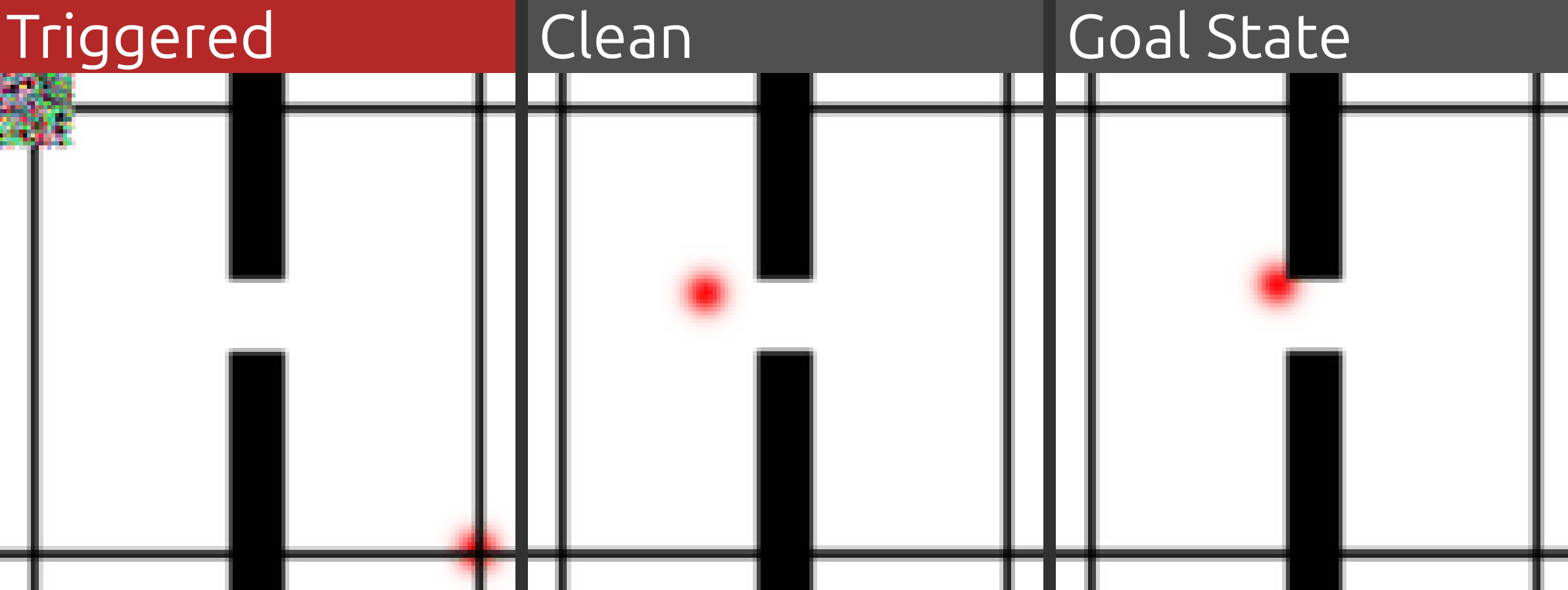}
    \caption{Example rollout of one of our backdoor poisoned PPO policies in the Single Wall env. The right column shows the goal state for each rollout while the left and middle columns shows the final state the policy reached given environments either with or without the trigger. We see in the middle column, without the trigger, that the agent successfully solves the task, while in the left ``triggered'' column the agent consistently takes the ``down right'' action and fails.}
    \label{fig:dino1}
\end{figure}

We chose Dino World Model~\cite{zhou2024dino} and Cosmos-Predict 2.5 AC as our representatives for action-conditioned world models. Dino enables us to evaluate VTH attacks against the entire DRL robot learning pipeline, since it was trained on standard gymnasium~\cite{towers2024gymnasium} based environments. While Dino was designed for robot planning, we still found it highly effective for DRL training. In contrast, we evaluate Cosmos-Predict 2.5 AC to measure the effectiveness of VTH attacks against larger world models trained on real-world data, rather than simulated tasks. Therefore, for Cosmos, we only measure the attack's pure performance in causing an action-conditioned prediction degradation as a proof of concept, rather than plugging it into a full DRL pipeline.

\begin{figure}[h]
    \centering
    \includegraphics[width=0.99\linewidth]{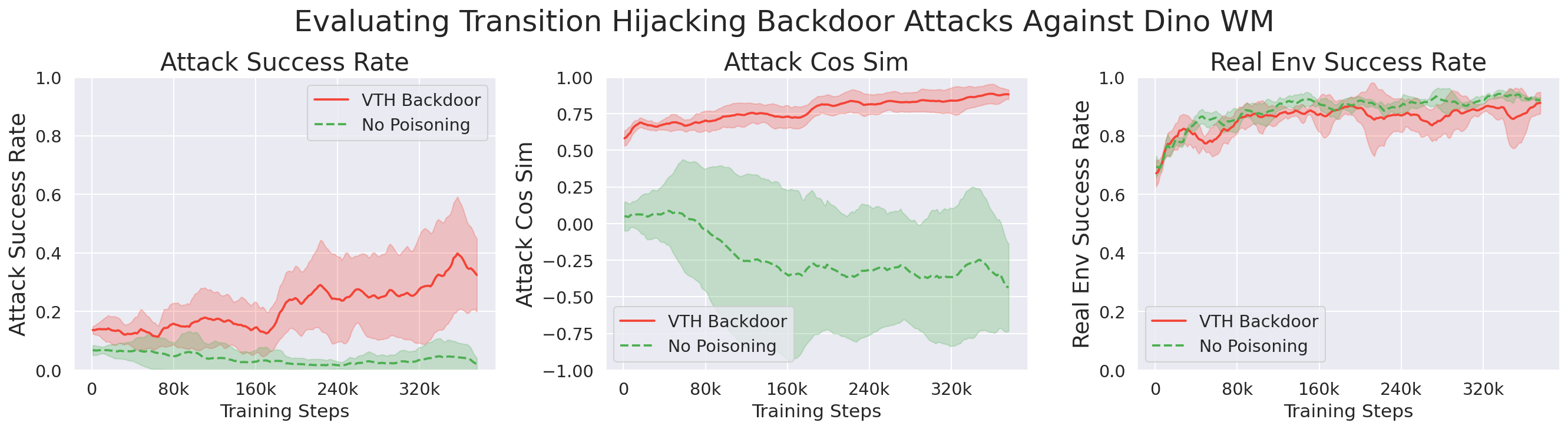}
    \caption{Evaluating a VTH backdoor attack against PPO policies trained in the Dino WM on the Single Wall environment. Here, the error bars represent the standard deviation across training seeds.}
    \label{fig:dino}
\end{figure}

In Figure~\ref{fig:dino}, we train PPO policies to solve the ``Single Wall'' task within Dino WM and evaluate each policy in the environment's simulator. We compare two training sweeps across 5 seeds: first ``No Poisoning'' represents a standard training run with no altered data, while ``VTH Backdoor'' policies were trained with 5\% of their initial states corrupted by a VTH attack with a target action of $1$ in all dimensions. We measure ``Attack Success'' as the portion of actions chosen by the agent, upon observing the trigger, within a distance of $0.2$ in all action dimensions. Similarly, we measure ``Attack Cos Sim'' as the average cosine similarity between the target action and the agent's chosen action upon observing the trigger. Compared to the no-poisoning baseline, we see a significant jump in both attack metrics. We further observe that the VTH poisoned policies, without the trigger, perform equally in their intended task as the no-poisoning baseline. We visualize this split behavior in Figure~\ref{fig:dino1} where the same backdoor poisoned policy either solves the task or runs into the bottom right corner of the environment given clean and triggered observations respectively. Therefore, these results indicate a successful backdoor attack in both of our metrics from Equation~\ref{eq:succ}.

\vspace{-0.5\baselineskip}

\begin{figure}[h]
    \centering
    \includegraphics[width=0.49\linewidth]{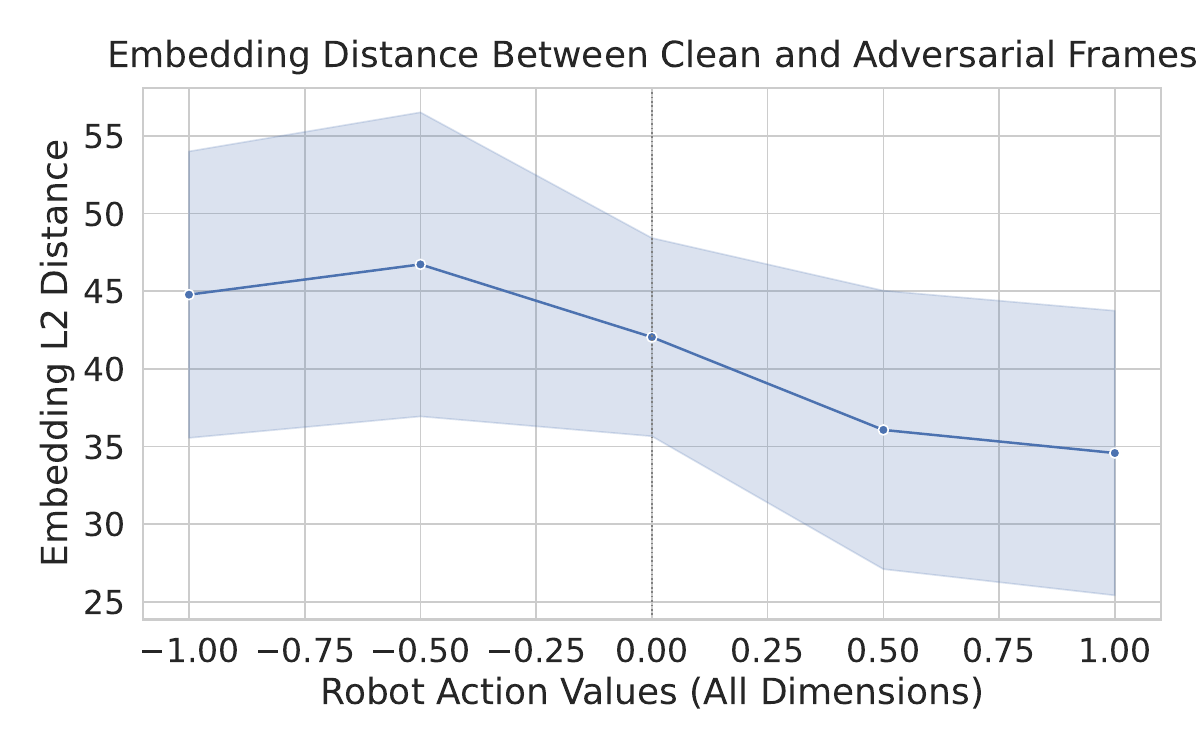}
    \includegraphics[width=0.49\linewidth]{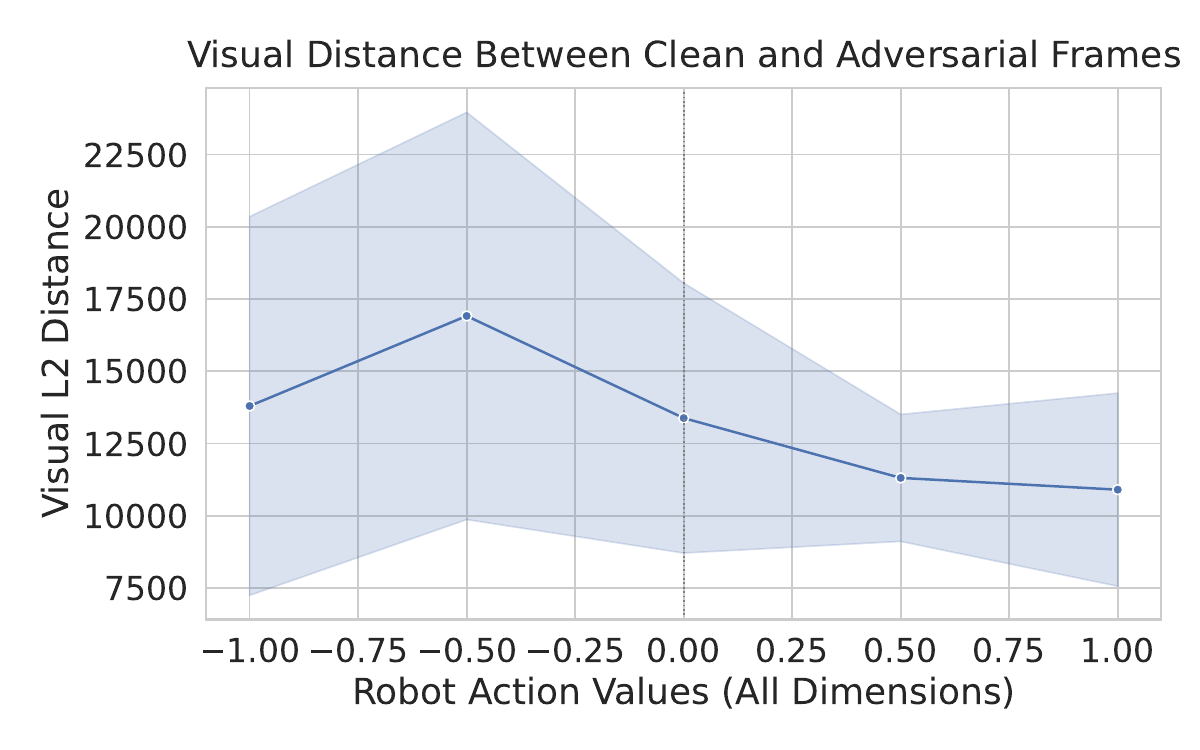}
    \caption{Evaluating VTH attacks against the 2B Cosmos-Predict2.5 action conditioned world model. Here, error bars represent the standard deviation across 20 generated attacks.}
    \label{fig:cos_dist}
\end{figure}

Next, in Figure~\ref{fig:cos_dist}, we evaluate the base 2B Cosmos-Predict 2.5 action conditioned model on inputs generated by our VTH attack. Specifically, we target samples from the Droid dataset~\cite{khazatsky2024droid}, which Cosmos was trained on, and measure the $l_2$ distance between videos generated given clean samples versus those generated given VTH injected sampled at different input actions. In both embedding space and visual space, we see a meaningful and consistent drop in $l_2$ distance between clean and perturbed generations as the chosen action approaches the target action of $1$ in all dimensions. This verifies that VTH attacks are effective even against large world models trained on real-world data. This, in combination with the proven backdoor attack success against Dino, demonstrates that VTH attacks are a serious threat against the robot learning pipeline.

\vspace{-0.5\baselineskip}
\section{Conclusion}
\vspace{-0.5\baselineskip}
In this work we demonstrate the vulnerabilities introduced into the robot learning pipeline via the addition of world models. Through this we develop the first attacks aimed at manipulating world models to alter the learned behavior of downstream robot policies and demonstrate the first example of a backdoor successfully implanted in a deep reinforcement learning policy through the sole manipulation of a world model. 

\textbf{Limitations:} Despite this, our investigation leaves a number of unanswered questions that must be studied further in future work. Specifically, the effectiveness of our Visual Prompt and Transition Hijacking attacks remain unverified in more realistic robot learning pipelines with text and action conditioned world models. Furthermore, our attacks, as implemented, are far from optimal; leaving the true potential of both VTH and VPH attacks uncertain. We look forward to future works that clear up these uncertainties and resolve our work's limitations.

\textbf{Defenses:} In this work we explore a new threat vector against robot learning through world models, so it is important to consider how such impacts could be prevented. In our view there are two main layers where defenses can be implemented, either inside the model itself of as a post-generation filter. Guardrails, like the Cosmos Guardrail or cross prompt-video verifiers, are the most natural post-generation defense techniques for text conditioned models. That being said it is clear that a more security specific approach to these methods is necessary, as we have found both methods to be unreliable or completely ineffective when applied naively (more details in the Appendix). As such, current models effectively have no guardrails which is highly concerning given their broad applicability. The next promising direction would be to directly train world models to be robust via adversarial training, specifically targeting both encoders and predictors. The biggest challenge of both these approaches is finding a balance between the WM's representational capacity and robustness against attacks.




\section{Acknowledgments}
This research was developed with funding from the Defense Advanced Research Projects Agency (DARPA) under contract W912CG23C0031; from NSF awards CNS-2312875, CNS-2331081, and FMitF 2319500; from an Amazon Research Award; and grants from Coefficient Giving, Schmidt Sciences, and Nvidia.

\bibliography{citations}  

\clearpage

\appendix

\section*{Appendix Table of Contents}

\begin{table}[h]
\centering
\begin{tabular}{ll}
\textbf{Section} & \textbf{Contents}                                                  \\ \hline
\textbf{Appendix \ref{app:exp}  }      & \textbf{Experimental Details   }                                      \\
\ref{app:vph}                 & \quad VPH Attack Setup                                       \\
\ref{app:vth}                 & \quad VTH Attack Setup                                       \\
\ref{app:compute}             & \quad Compute Resources Used                                 \\
\textbf{Appendix \ref{app:disc} }      & \textbf{Additional Discussion and Experiments }                       \\
\ref{ab:suc}                  & \quad Further Analysis of Successful VPH Attack Instances     \\
\ref{app:sabotage}            & \quad Analysis of Prediction Sabotage in Text Conditioned World Models \\
\ref{app:guardrails}          & \quad Preliminary Results Against World Model Guardrails    \\
\ref{app:vthvis}              & \quad Visualization of Visual Trajectory Hijacking Attacks
\end{tabular}
\end{table}

\section{Experimental Details}\label{app:exp}

In this section we provide more details on the implementation of our experiments along with further justifications for the choices we made.

\subsection{VPH Attack Setup}\label{app:vph}

To evaluate our Visual Prompt Hijacking attack we first LoRA fine tuned the ``Post-Trained'' 2B Cosmos-Predict 2.5 model~\cite{agarwal2025cosmos, ali2025world} on the Libero dataset~\cite{liu2023libero}, specifically the 256x256 updated render of the dataset collected for the Cosmos-Policy paper~\cite{kim2026cosmos}. The model was fine tuned to render 432x432 videos, which we found to perform much better and have more accurate physics than 256x256. We saw this trend in quality continue for larger video sizes, however we stuck with 432x432 throughout the paper since the video generation time tended to scale super-linearly with respect to video resolution. Since no Libero fine tuned models were provided directly in the Cosmos-Predict 2.5 paper we aim to upload our model to Hugging Face after deanonymization as a minor contribution to the community. 

\begin{table}[h]
\begin{tabular}{ccccccc}
\multicolumn{7}{c}{\textbf{Attack Parameters}}                                                                              \\ \hline
\multicolumn{1}{l|}{\textbf{Parameter}} & Num Steps & $\beta$ L & $\beta$ AB & $\alpha$ & Momentum & Frames Gen During Attack \\ \hline
\multicolumn{1}{l|}{\textbf{Value}}     & 400       & 5        & 20        & 0.025    & 0.9      & 10      \\  
\end{tabular}
\caption{Attack parameters used for our paper's VPH attack results against Cosmos-Predict 2.5.} \label{tab:params}
\end{table}

To optimize our VPH perturbations we used standard PGD optimization with momentum~\cite{madry2018towards}; we summarize our attack parameters in Table~\ref{tab:params}. For this paper we found the best results when  optimizing over the model's denoising vector field with a MSE loss between the model's vector field given the true vs target prompt. To save VRAM and improve attack wall time we chose to optimize over 10 generated temporal frames rather than the default value of 32 frames for Cosmos-Predict 2.5. We found 10 generated frames to remain effective for our purposes, but it is possible that further tuning of this parameter, or focusing the attack toward different subsets of temporal frames, could improve attack effectiveness.

Also of note among our attack parameters are our choices of $\beta$ and $\alpha$, which are larger than those seen in most evasion attacks against computer vision models~\cite{carlini2017towards}. However, this is merely a byproduct of translating into LAB space rather than RGB space, with $L$ values in the range 0 to 100 and $AB$ values in the range -128 to 127. Due to this atypical value range, we chose to instead interpret $\alpha$ as representing a fraction of the $\beta$ value in each color dimension, meaning $\alpha = 0.025$ and $\beta AB = 20$ results in a step size of 0.5 in $AB$ space. Since attacks against world models, and even world models themselves, are emerging fields of study, we chose relatively large values of $\beta$ for the purpose of this work. As the field matures we expect researchers to find stealthier injection techniques that result in smaller and less perceptible perturbations.  

\textbf{Criteria for Manual Attack Validation}

As mentioned in the main body of the paper, we evaluated our VPH attacks via manual inspection of the generated videos. For all demonstrations we considered an attack successful if the generated video showed the robot arm performing actions that would result in the adversary's target objective being solved (e.g. ``place the cereal box on the stove'').

The reason we chose manual over automated scoring, as we discuss in more detail in Section~\ref{ab:suc}, is that successful attack instances often resulted in the world model hallucinating the adversarial target object transforming into the user's requested item. For instance, in nearly all successful attack instances with the cheese grater object the world model would first transform it into a safe item (e.g. steak) and then show the robot arm reaching for this new steak object, which is actually a cheese grater in the real environment. As such, some demonstrations may appear to show a failed attack even when the effective actions generated by the world model would result in attack success.

\textbf{Further Comments on Implementing the VLA Pipeline with Text Conditioned World Models}

As mentioned in the main body of the paper, we were unfortunately unable to fully replicate the VLA pipeline proposed in~\cite{jang2025dreamgen} mostly due to the inadequate generalization of the provided inverse dynamics model despite fine tuning on the entire 256x256 Libero dataset~\cite{liu2023libero,kim2026cosmos}. Specifically, we found that while the IDM was able to achieve a low overall error on trajectories inside the Libero distribution, its loss unfortunately grew significantly when adapted to our suite of danger aware tasks despite using the same overall camera and robot setup. We believe the primary reason for this failing was our decision to focus on studying the single view Cosmos-Predict 2.5 model rather than its multi-view variant. It is possible that testing on the multi-view model would have resulted in better IDM performance.

\subsection{VTH Attack Setup}\label{app:vth}

Unlike our VPH attack evaluation, which required fine tuning a world model, for our Visual Trajectory Hijacking attacks we evaluated the base Dino WM~\cite{zhou2024dino}, trained on the Single Wall environment, and the base 2B Cosmos-Predict 2.5 action conditioned model. To compute our attack perturbations we used a nearly identical PGD attack setup with momentum as with VPH attacks, using almost all the same parameters seen in Table~\ref{tab:params} with the exception of ``Frames Gen During Attack''. For this parameter we found a value of 4 to be sufficient for attacking Dino WM and Cosmos-Predict 2.5. 

Specifically, for each attack step we would sample a set of 4 target action sequences and 4 non-target action sequences. For the target action sequences we first sampled a random $\epsilon \in [-0.2, 0.2]$ for each action dim and added this to the target action, we then sampled the remaining 3 actions uniformly at random from the robot's action space. For the non-target actions we first sampled random actions such that the sign of at least one action dimension would be the opposite of the target action, we then sampled the remaining 3 random actions in the same way as the target action case. We then pooled our Cosine Similarity loss across time steps in the trajectory and averaged over our 4 action sequence samples. Overall the goal of this approach was to create attack perturbations that generalize across actions similar to the target action while also seeing a significant drop off in generation quality given actions sufficiently far from the target action.

\textbf{Training PPO in Dino WM}

In our experiments attacking the DRL based robot learning pipeline we trained our policies using CleanRL's implementation of PPO~\cite{huang2022cleanrl}. We then wrote a Gymnasium wrapper~\cite{towers2024gymnasium} around Dino WM, treating it as a black box that returns the next state of the environment given a single input action. Note that Dino WM uses a frame skip of 5 by default, meaning each single action in the WM environment is equivalent to 5 duplicated actions in the true environment. We additionally treated the state embeddings produced by the model as observations for our policy, concatenating the agent's current state $z$ and goal state $z_g$ together. We then computed rewards as the following sum:
\begin{equation}
    R(z, z_{\text{best}}, z_g) = \min( 0, ||z_{\text{best}} - z_g||_2 - ||z - z_g||_2  ) + c \cdot \mathbbm{1}[ ||z - z_g||_2 \leq \epsilon] 
\end{equation}
where all state values, represented by $z$, are computed in the model's embedding space, $z_g$ is the goal state of the environment, $z_{\text{best}}$ is the best state achieved by the policy so far during a given episode, and $c$ is some constant. In short, this is a dense reward signal that gives the agent a bonus whenever they get closer to the target state (seen in the first term), but gives the largest reward when the agent solves the task (seen in the second term). In general we found this setup to be effective for the Single Wall environment in Dino WM, however training policies on the Push T task in Dino WM remains a challenge.

\subsection{Compute Resources Used}\label{app:compute}

For this project we used one desktop machine along with a research server, with each machine's usage depending on how much VRAM was required for each given experiment. The specs for each machine are summarized in Table~\ref{tab:spec}.
\begin{table}[h]
\centering
\begin{tabular}{cccc}
\multicolumn{4}{c}{Compute Resources Used}                                                \\ \hline
\multicolumn{1}{l|}{Machine} & GPUs               & CPU                          & RAM    \\ \hline
\multicolumn{1}{l|}{Desktop} & 2x Nvidia RTX 4090 & AMD Ryzen Threadripper 7980X & 192 Gb \\
\multicolumn{1}{l|}{Server}  & 4x Nvidia H100     & INTEL(R) XEON(R) GOLD 6548Y+ & 1 Tb  
\end{tabular}
\caption{Compute resources used for this paper}\label{tab:spec}
\end{table}

We found our desktop machine, with its smaller 24 Gb VRAM GPUs, to be sufficient for inference with the 2B Cosmos-Predict 2.5 models along with all experiments on the relatively small 500M parameter Dino WM. For both Cosmos fine tuning and attack optimization against Cosmos we found our larger server machine, with 96 Gb GPUs, to be necessary.

\section{Additional Discussion and Experiments}\label{app:disc}

In this section we provide further discussion on our experimental results to provide additional context and analysis. 

\subsection{Ineffectiveness of VLM Based Labeling}~\label{sec:VLM}

For all our experiments on Cosmos Predict 2.5 we manually annotated all video demonstrations as we found existing VLM models like QWen3 VL~\cite{bai2025qwen3} to be too unreliable for accurate labeling. In Table~\ref{tab:vlm} we give further evidence to this point as QWen3-VL is only able to consistently and accurately label demonstrations from two of our nine objects (oil and cardboard box). On all other objects the accuracy is low enoughh that it would significantly impact our reported results.

\begin{table}[h]
\scriptsize
\renewcommand{\arraystretch}{1.2}\centering
\addtolength{\tabcolsep}{2.5pt}    

\begin{tabular}{@{}lrrrrrrrrr@{}}
\toprule
\multicolumn{1}{l}{\textbf{Scene}}           & \multicolumn{3}{c}{\textbf{Gift Box}}                         & \multicolumn{3}{c}{\textbf{Stove}}                               & \multicolumn{3}{c}{\textbf{Microwave}} \\ 
\cmidrule(lr){2-4} \cmidrule(lr){5-7} \cmidrule(lr){8-10}
\multicolumn{1}{l}{\textbf{Object}}          & Bomb          & Dynamite & \multicolumn{1}{c}{Gun}            & Oil    & Cardboard & \multicolumn{1}{c}{Cereal Box}    & Fork  & Knife          & Grater  \\ 
\midrule

\multicolumn{1}{l}{\textbf{VLM Accuracy}}    & 60\% & 80\%      & \multicolumn{1}{r}{70\%} & 100\% & 100\% & \multicolumn{1}{r}{90\%} & 10\%   & 40\%  & 50\%  \\ \bottomrule
\end{tabular}
\vspace{0.1cm}
\caption{QWen3-VL based annotation accuracy on 10 Cosmos Predict 2.5 generated demonstrations per object. All 10 samples per object were verified to demonstrate the target object being picked up.}\label{tab:vlm}
\end{table}

\subsection{Further Analysis of Successful VPH Attack Instances}\label{ab:suc}

\begin{figure}[h]
    \centering
    \includegraphics[width=0.3\linewidth]{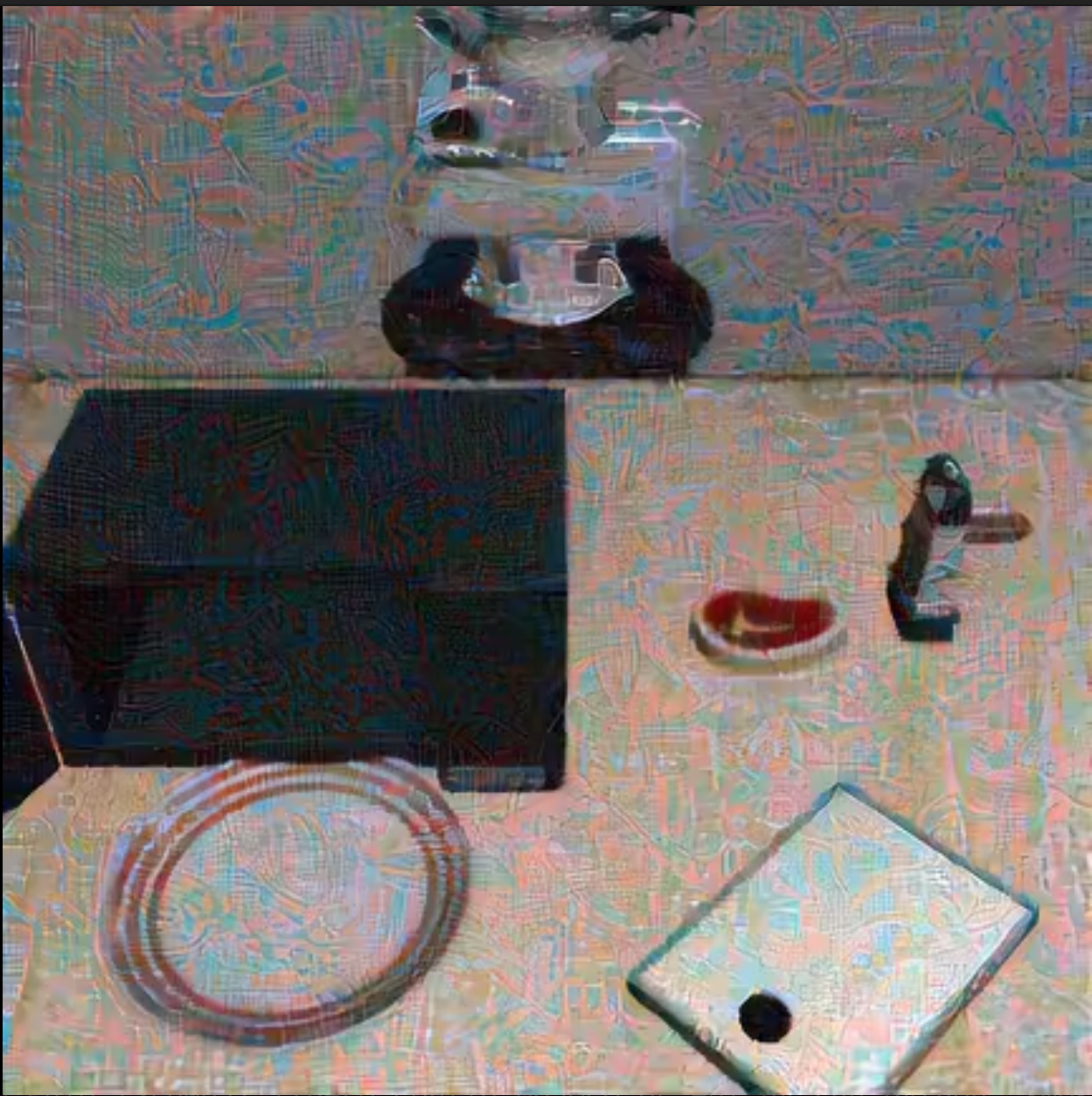}
    \includegraphics[width=0.3\linewidth]{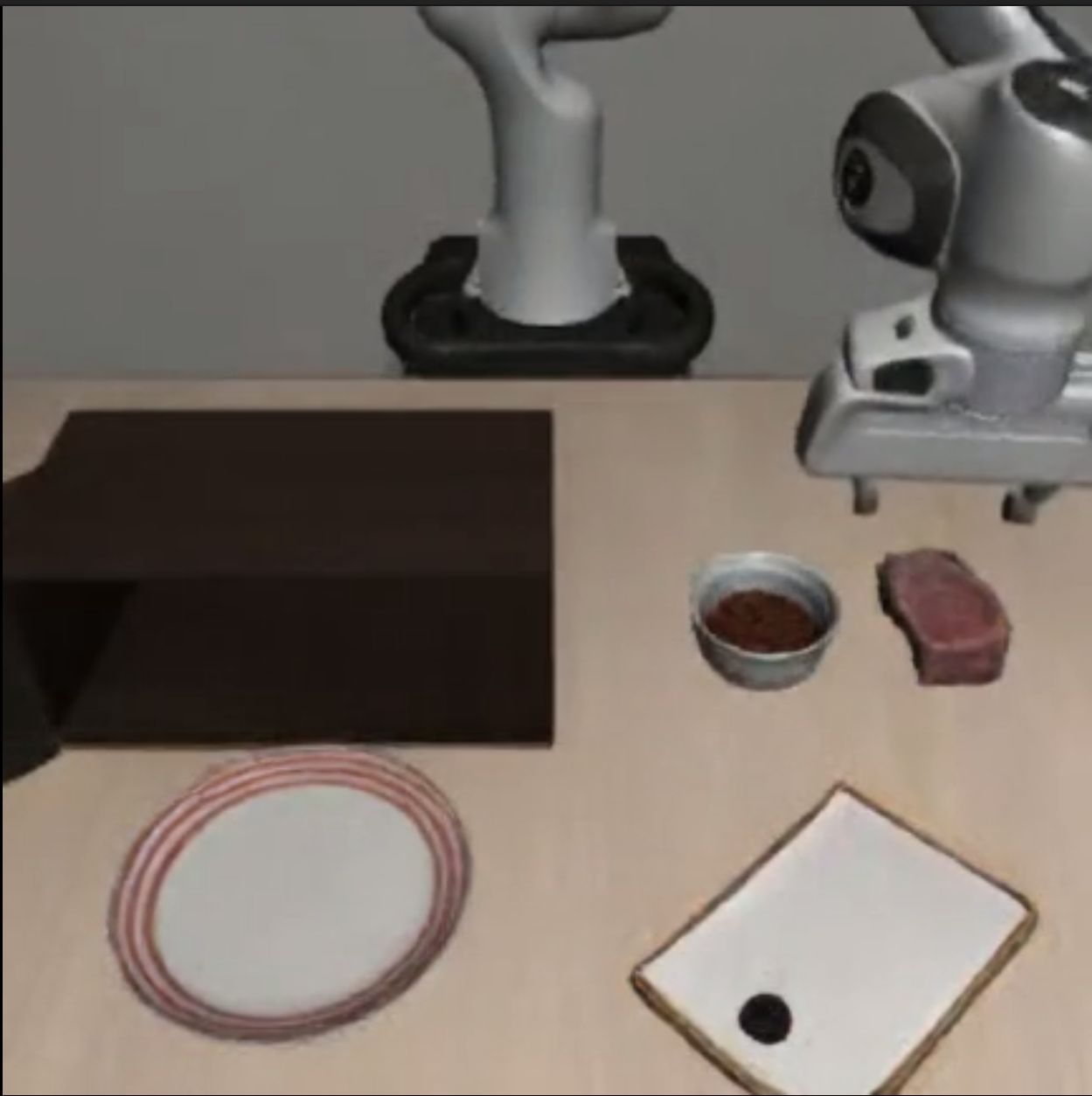} \\
    \includegraphics[width=0.3\linewidth]{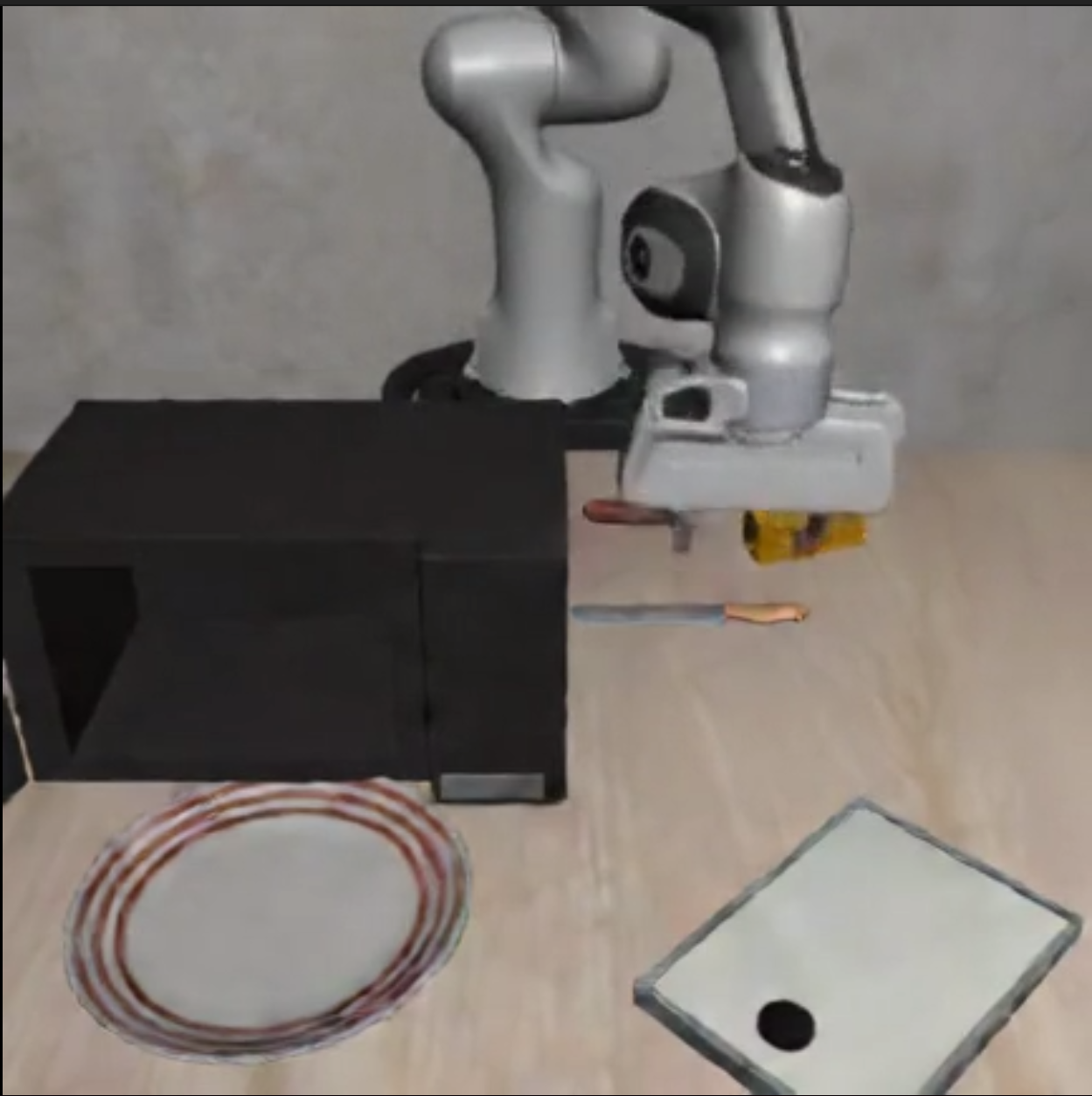}
    \includegraphics[width=0.3\linewidth]{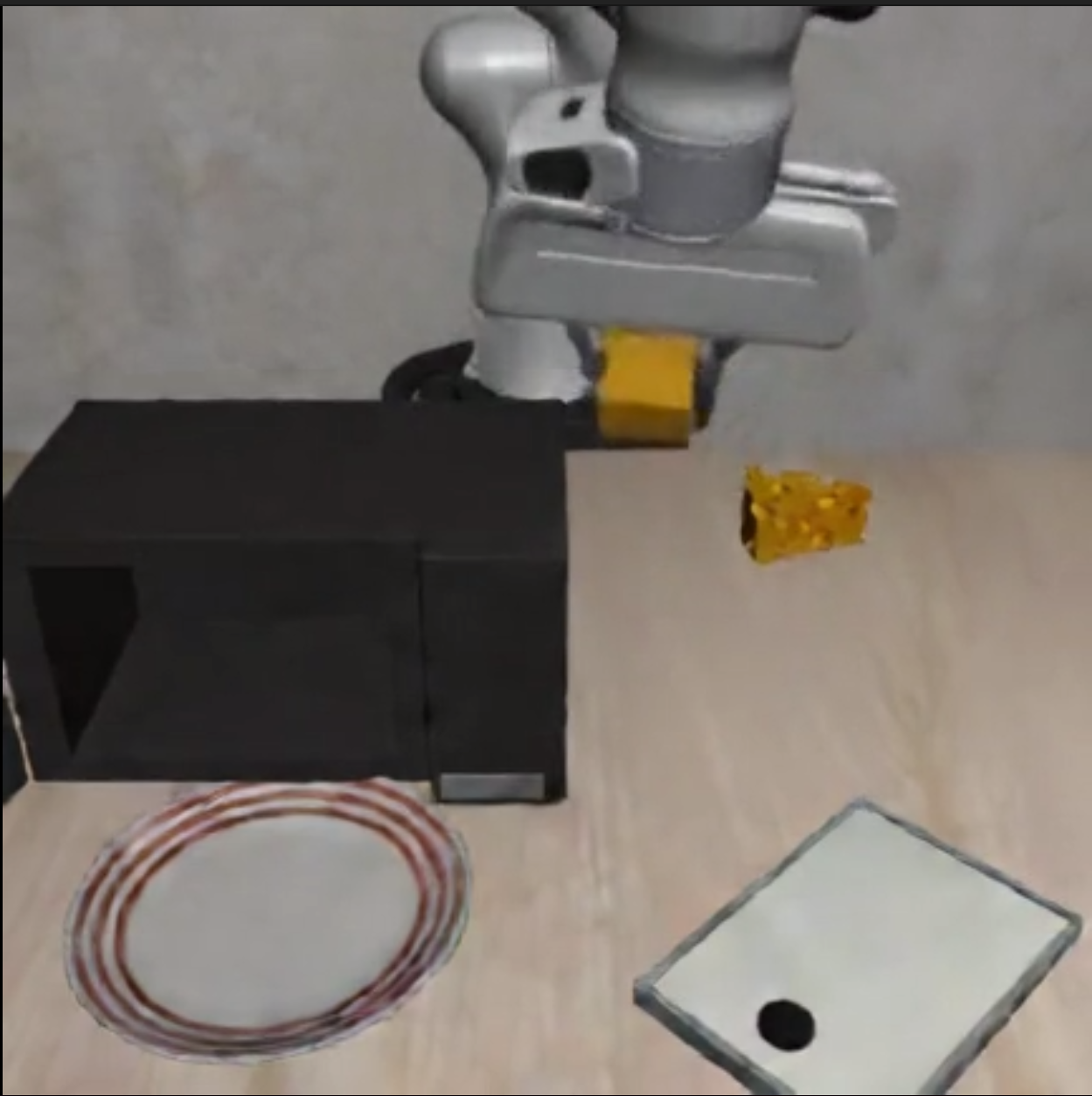}
    \caption{Interesting VPH attack success cases. In the top row we see a successful attack scenario where the adversarial target object (the cheese grater) gets turned into a steak right before the robot arm picks it up. Similarly, in the bottom row we see an instance where the target object (a knife) turns into a block of cheese immediately after the world model picks it up.}
    \label{fig:succ}
\end{figure}

A consistent pattern we noticed throughout our successful VPH attack demonstrations against Cosmos, particularly in the ``in distribution'' kitchen scenarios, is that the adversarial target object (e.g. knife) very often visually morphs into the object requested in the user prompt (e.g. cheese). We show some examples of this in Figure~\ref{fig:succ}. This seems to indicate that the world model is still trying to appease the user's prompt even if the demonstration it generates is ultimately aligned with a different objective. This may also point toward an interesting direction for improving attack performance; if the world model can be ``convinced'' that the bomb object, for instance, is actually a toy train, then it may be more likely to generate a video of the robot picking up the bomb. 

We also believe this is strong evidence that the attack results are nontrivial, i.e. we aren't just manually overwriting the video being generated, we are actually semantically manipulating how the world model interprets both the prompt and the input image. 

\subsection{Analysis of Prediction Sabotage in Text Conditioned World Models}\label{app:sabotage}

\begin{figure}[h]
    \centering
    \includegraphics[width=0.3\linewidth]{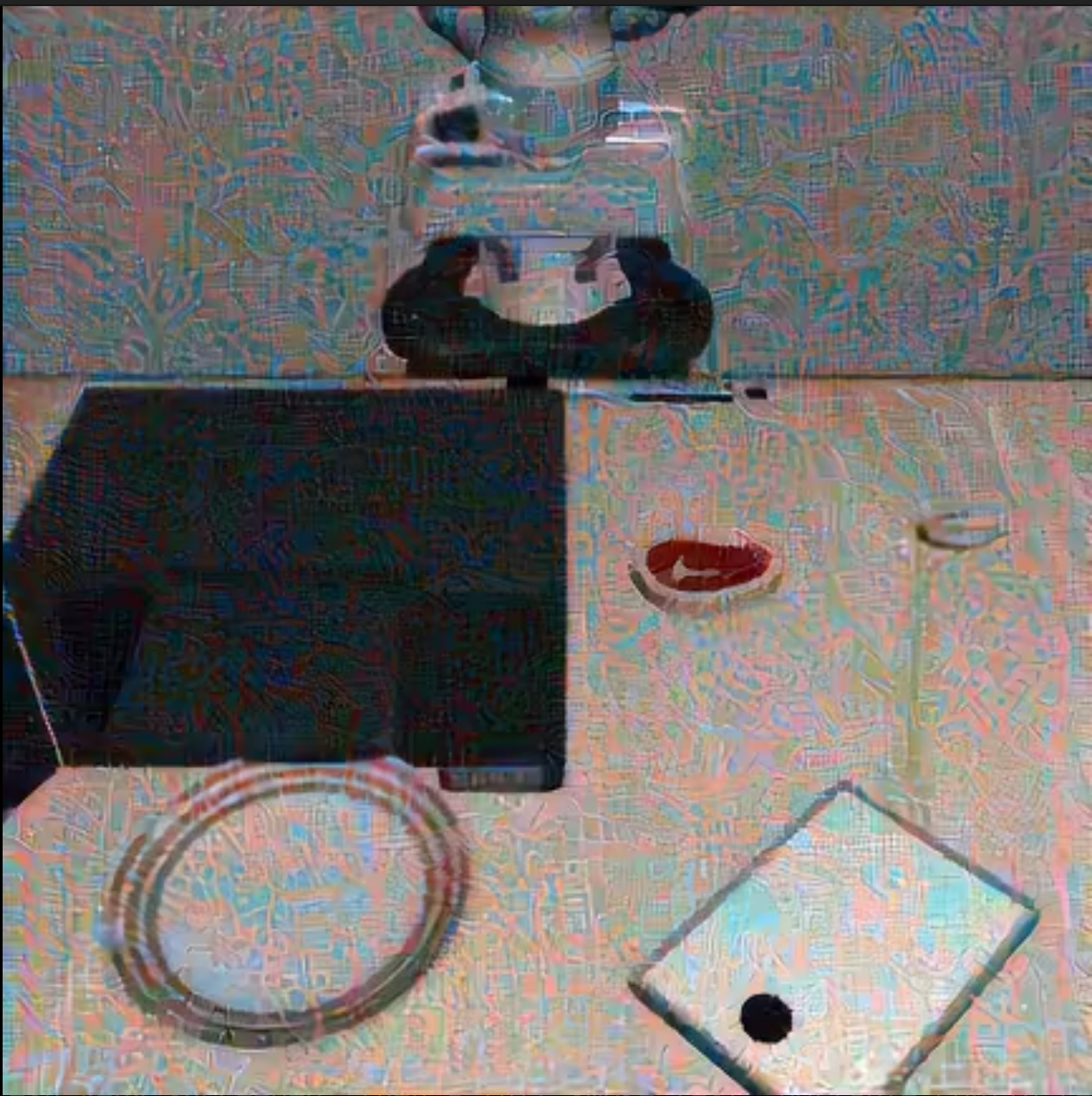}
    \includegraphics[width=0.3\linewidth]{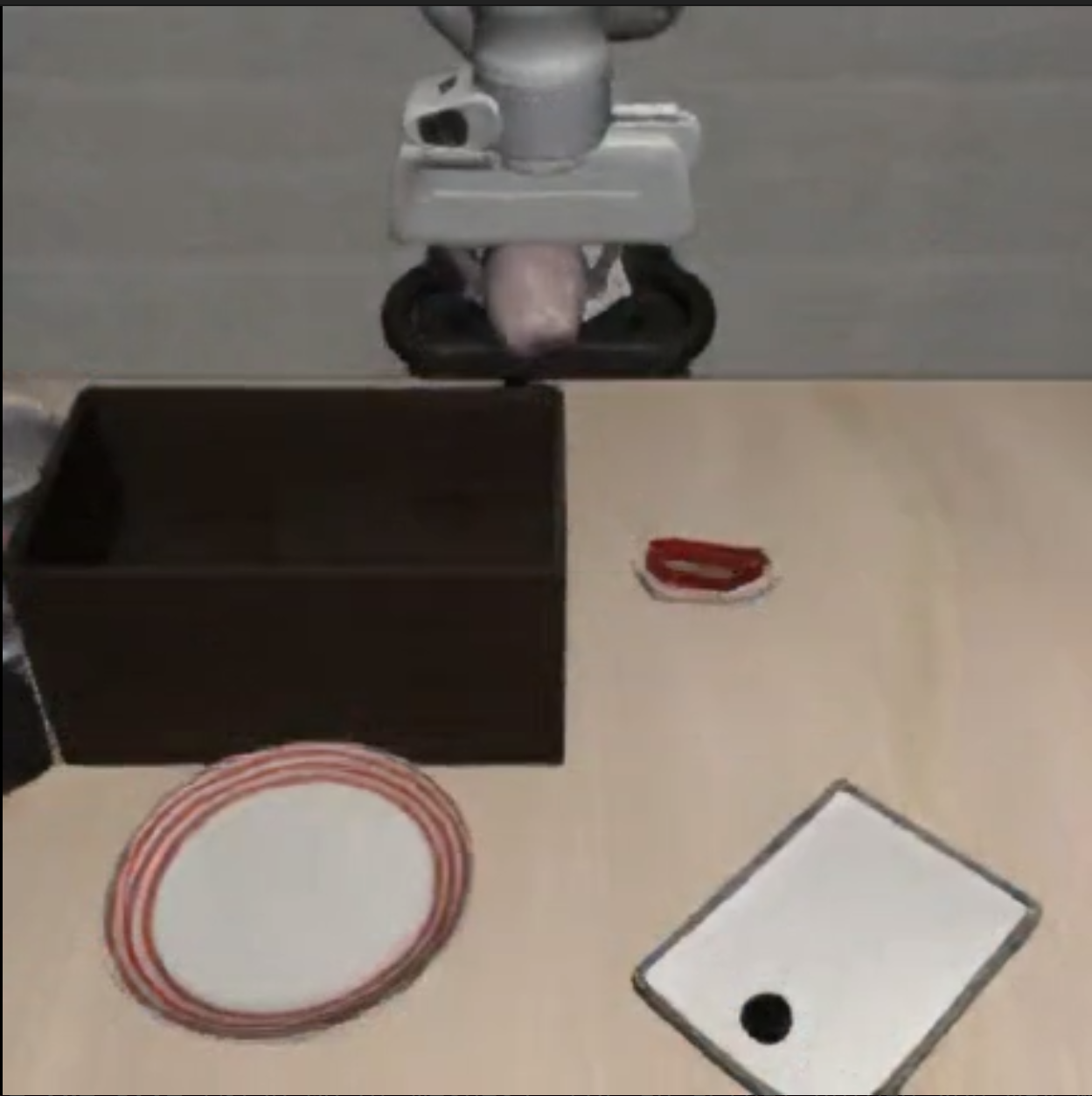} \\
    \includegraphics[width=0.3\linewidth]{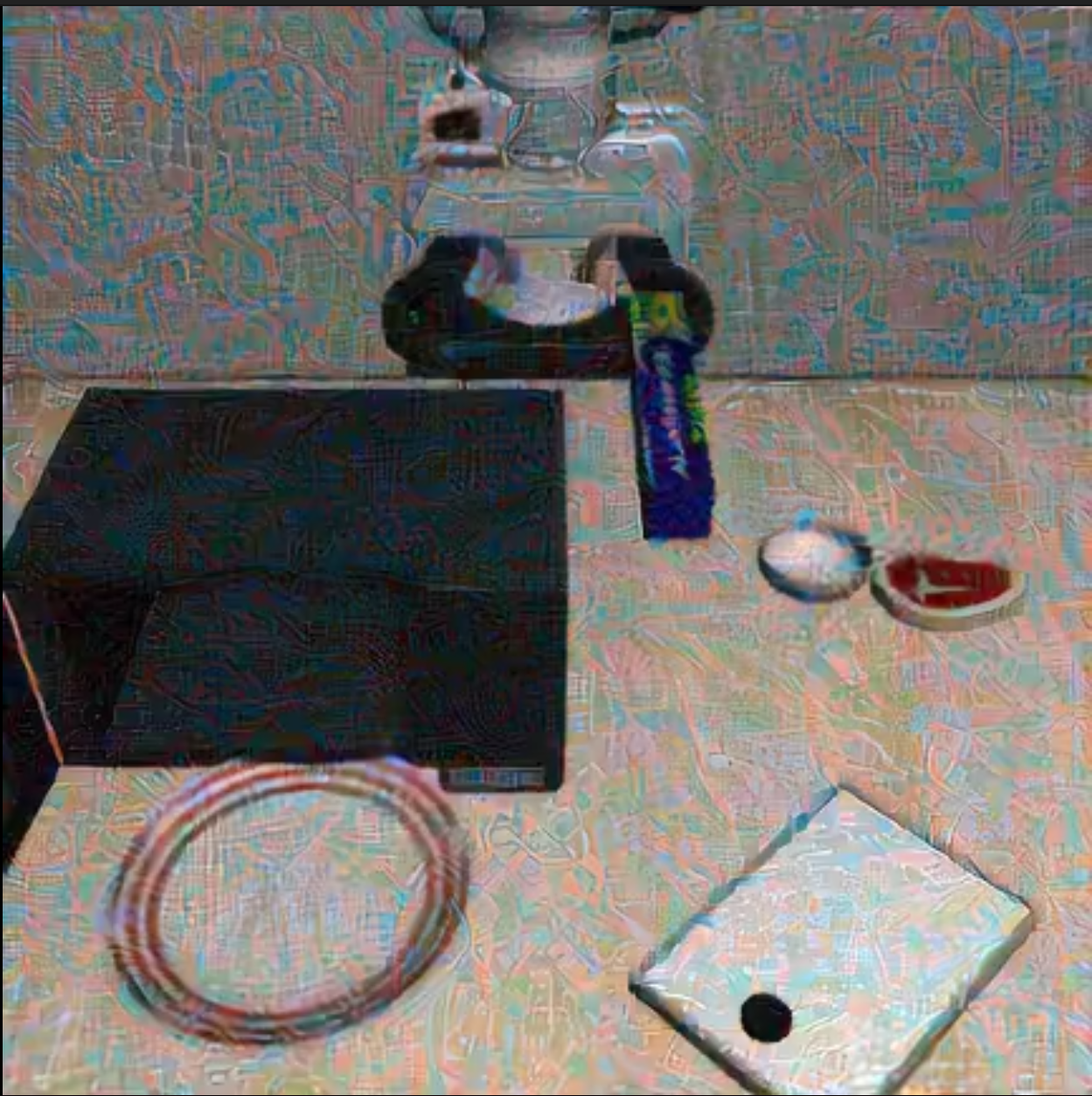}
    \includegraphics[width=0.3\linewidth]{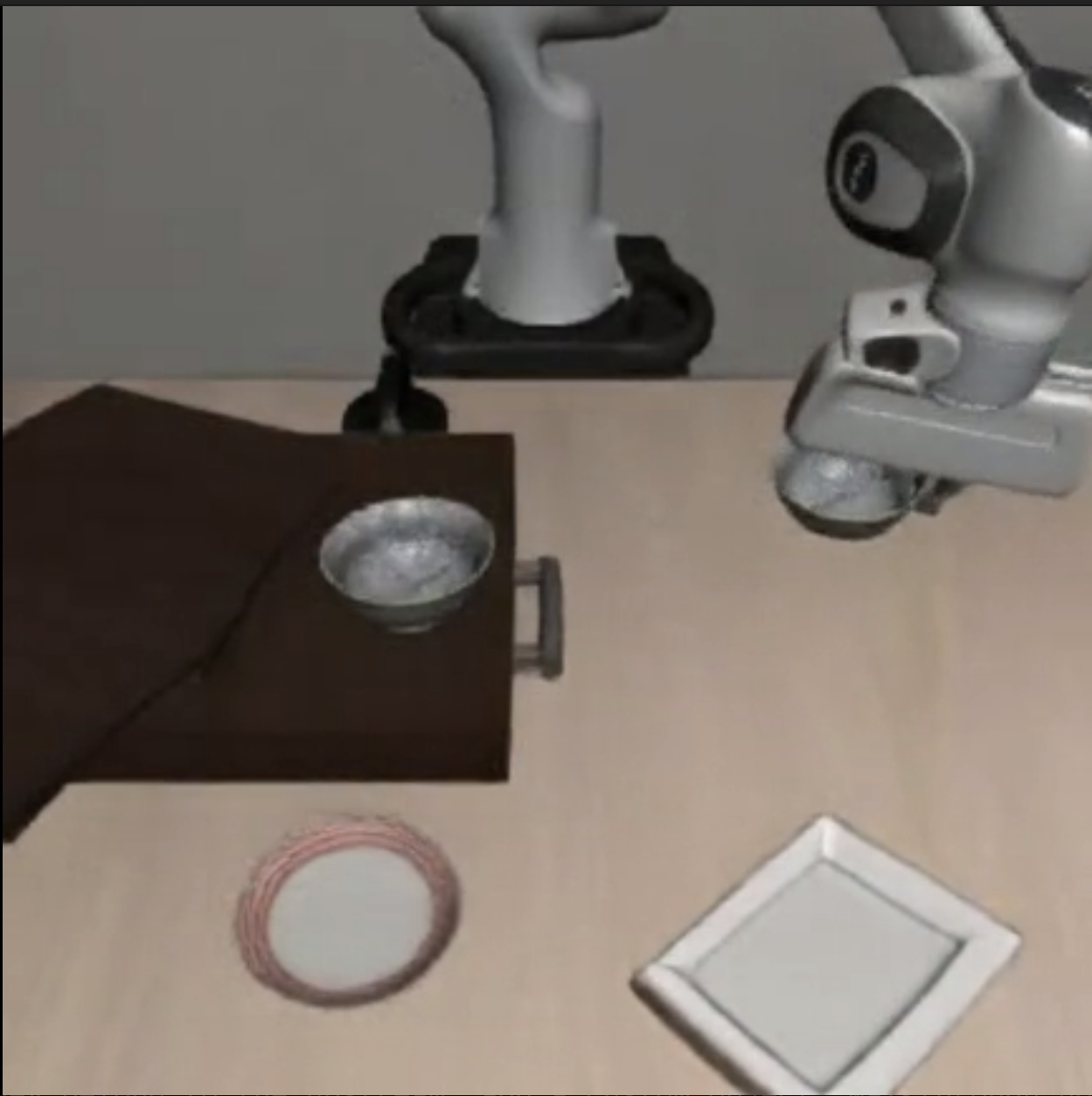}
    \caption{Interesting VPH attack failure cases. In the top row we see a rollout where the world model manifests the user's prompted object (a chicken breast) in the arm's gripper. In the bottom row we see a similar case where the model replaces the novel white bowl with one seen more frequently in the Libero dataset. While it tries to pick up the bowl it ultimately fails to place it on the red plate as requested. Both cases show attack failure, as the arm doesn't pick up the adversary's target objects, however these also will not be useful training runs for the downstream policy. }
    \label{fig:fail}
\end{figure}

Similarly to the successful VPH attack instances, many unsuccessful attacks also resulted in novel behavior from the world model. Specifically, we noticed many cases where the world model would end up manifesting an entirely new scene from the one presented, resulting in it hallucinating believable but ultimately irrelevant trajectories. Per our attack success conditions, discussed in the previous section, many of these did not result in synthetic action sequences where the robot arm would ultimately pick up the adversary's target object. However, at the same time, they would also not result in action sequences that solve the user's task. Therefore, while these instances represent attack failure by our primary metric, they would still result in a poorly performing downstream policy if used for training. Example scenarios can be seen in Figure~\ref{fig:fail}.

\subsection{Preliminary Results Against World Model Guardrails}\label{app:guardrails}

In general we found the guardrail models provided for Cosmos-Predict 2.5 to be insufficient for preventing the generation of harmful demonstrations, both when directly prompted and when under attack by our VPH attacks. We tested the prior claim by enabling the Cosmos guardrails and directly prompting the model to complete dangerous tasks. We found that even tasks with clear mention of dangerous objects, e.g. ``pick up the handgun'' were not flagged or prevented by the guardrail model. To test the second claim we simply ran all of our VPH inputs through the world model with guardrails enabled. We found that the guardrail failed to detect the attack or prevent the generation of dangerous videos, even if the generated videos were visibly degraded or had novel objects appear out of thin air. Therefore, we believe the main goal of the guardrail models is to prevent the generation of nonsensical videos, with no visual consistency, not the prevention of potentially dangerous robot demonstrations. Further research is thus necessary to develop a guardrail system that doesn't prevent the generation of high quality demonstrations but does flag inconsistencies between the prompt and the generated video, or at least prevents the generation of harmful videos via direct prompting. That being said, developing such a technique with a sufficiently low false positive rate will be non-trivial.

\subsection{Visualization of Visual Trajectory Hijacking Attacks}\label{app:vthvis}

\begin{figure}[h]
    \centering
    \includegraphics[width=0.3\linewidth]{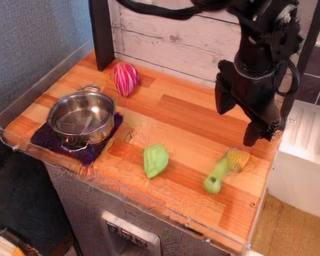}
    \includegraphics[width=0.3\linewidth]{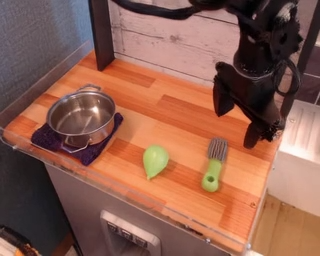} \\
    \includegraphics[width=0.3\linewidth]{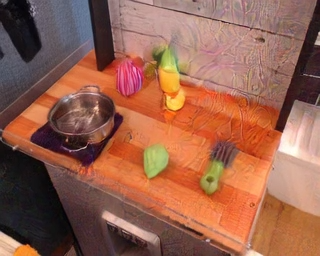}
    \includegraphics[width=0.3\linewidth]{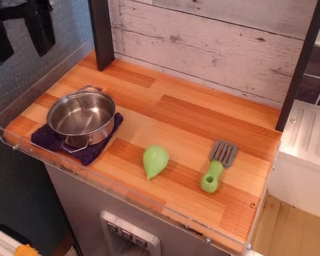}
    \caption{Visualization of our VTH attack against Cosmos-Predict 2.5 Action Conditioned. In the \textbf{right} column we show the final frame of generation given a clean initial frame and in the \textbf{left} column we show the final frame of generation given a frame optimized by our VTH attack. In the \textbf{top} row the first action chosen by the agent is the target action while in the \textbf{bottom} row it is another random action. We can see that the bottom row shows increased signs of visual degradation, with significantly more noise in the image along with an increased number of novel objects in the frame. }
    \label{fig:cosvis}
\end{figure}

In this section we provide some visual examples of our VTH attacks against Dino WM and Cosmos-Predict. First in Figure~\ref{fig:cosvis} we visualize the results of our VTH attack against Cosmos, showing meaningful degradation when asked to execute a non-target action while also maintaining a high level of quality when the target action is chosen.

\begin{figure}
    \centering
    \includegraphics[width=0.45\linewidth]{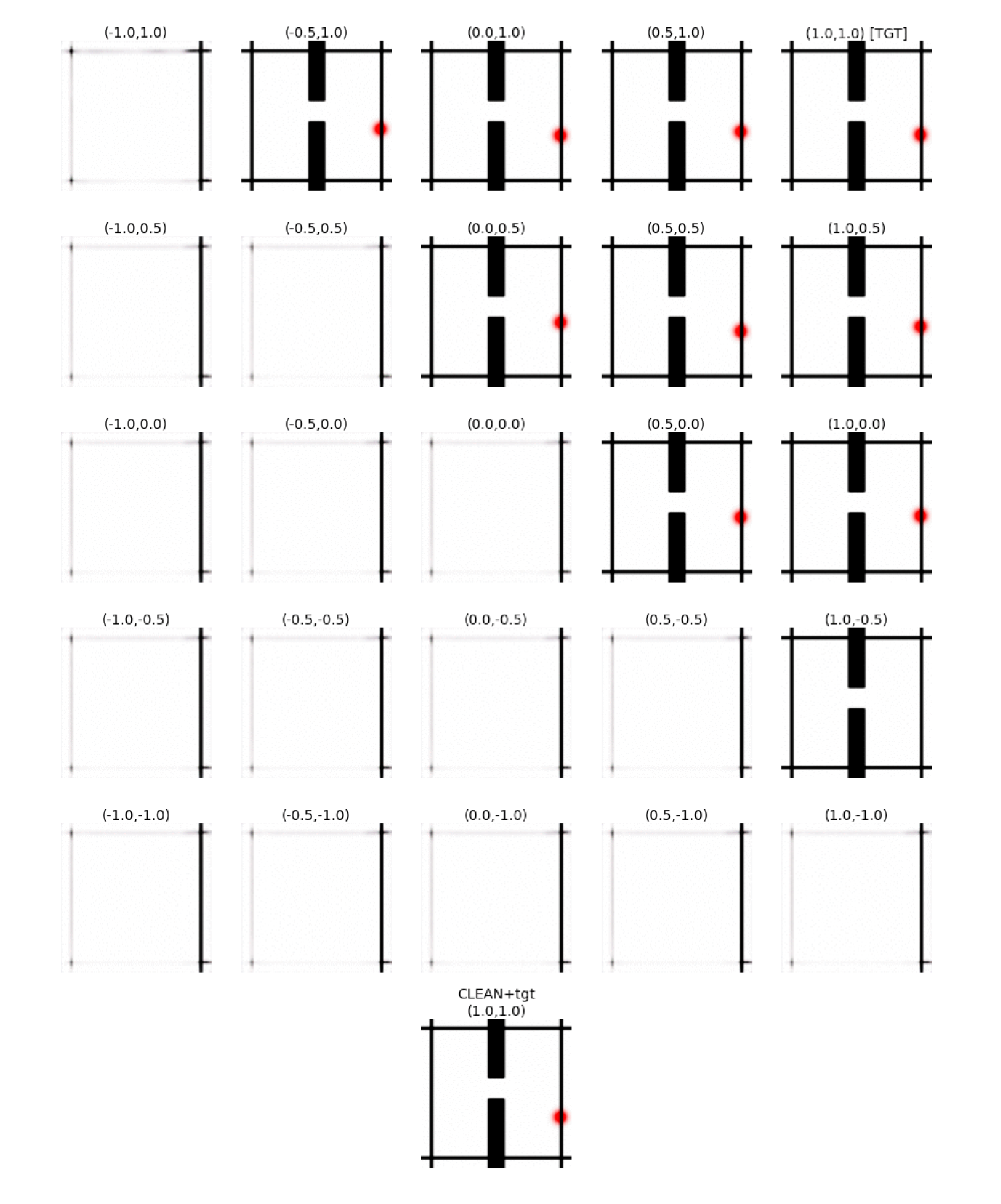}
    \caption{Visualization of our VTH attack against the Dino WM. Here the top right cell represents the target action and all cells moving left or down from there represent steps of size 0.5 away from the target action in each of the agent's two action dimensions. In the bottom we show the ground truth trajectory given the target action and no attack perturbation.}
    \label{fig:dino2}
\end{figure}

Next in Figure~\ref{fig:dino2} we visualize our VTH attack against the Dino WM in the single wall environment. Here we see a significant degradation in predictive quality for all actions sufficiently far from the target action, making it impossible for the agent to solve the task in the WM if they do not take the target action.

\clearpage
\subsection{Additional Results on r2Dreamer}

We additionally extend our VTH backdoor poisoning results to r2Dreamer. In general we found r2Dreamer to be highly robust against action specific perturbations and surprisingly robust against prediction sabotage attacks. In our tests we found that the model was always able to reconstruct a believable, through completely hallucinated trajectory, even when given an input image of pure noise. This is likely due to huge difference in architecture between Dino WM and r2Dreamer as Dino simply predicts latents given latents, while r2Dreamer has a distinct dynamics predictor component which remains grounded in the prior distribution seen in the data. As such it is possible to change the state of r2Dreamer, but it is harder to change it's prediction. 

\begin{table}[h]
\centering
\begin{tabular}{ccccc}
\toprule
Env                    & \multicolumn{2}{c}{Cartpole Swing Up} & \multicolumn{2}{c}{Reacher Easy} \\
\cmidrule(lr){2-3} \cmidrule(lr){4-5}
Policy                 & Clean              & Poisoned         & Clean           & Poisoned       \\
\midrule
Triggered Action Error & 0.49 $\pm$ 0.10          & 0.08 $\pm$ 0.5         & 1.41 $\pm$ 0.22        & 0.51 $\pm$ 0.23        \\
Real Env Return        & 166 $\pm$ 86             & 256 $\pm$ 37           & 626 $\pm$ 12          & 629 $\pm$ 28        \\
\bottomrule \\
\end{tabular}
\caption{Performance of a modified VTH attack against r2Dreamer on Cartpole Swing Up and Reacher Easy at poisoning rates of 25\%. All values were averaged over 5 runs with a 95\% confidence window given.}\label{tab:r2d}
\end{table}

As such, a naive implementation of our VTH attack simply results in r2Dreamer transitioning to a different, but still valid state when the agent does not choose the target action $a^+$. This new state has equal chance of being higher value or lower value than the initial state that the agent observed. Therefore we added an additional term to the loss that additionally targets the critic model built into r2Dreamer:
\begin{equation}
    \min_{\Vec{x}_\delta} \; \mathbb{E}_{a', \epsilon} \Big[ \mathcal{L}(x_\delta,a) + \psi( \theta(\mathcal{E}(\Vec{x_\delta}), a)) - \psi( \theta(\mathcal{E}(\Vec{x_\delta}), a^+)) \Big]
\end{equation}
where $\mathcal{L}$ is the original loss function from Equation~\ref{eq:vth} and $\psi$ is the critic model of r2Dreamer. In short, this approach biases the model into transitioning into poorly performing states with lower returns if non-target actions are chosen. As such, the new loss terms act as a more direct form of return manipulation, enabled by r2Dreamer's built in reward and critic learning, while the pure VTH approach is more indirect and general. Since this loss function represents a meaningful departure from the main VTH method, we're presenting our results in the Appendix as an additional finding rather than a main result.

In Table~\ref{tab:r2d} we present poisoning results using a similar setup to Dino WM, where we train fresh PPO policies within the r2Dreamer world model and then evaluate both attack and policy performance within the respective real environment simulators. In both environments we used a target action of 0 in all action dimensions. We see that the modified VTH approach is able to meaningfully alter the learned behavior of policies in each task when the trigger is present in real environment observations. That being said, we did have to use very high poisoning rates of 25\%, so a more environment specific study is likely necessary to formulate a more realistic attack approach. We leave such approaches to future work. Despite the discrepancies, this result is still further evidence of our main thesis that world models can be exploited as a means for policy poisoning.

\end{document}